\documentclass[letterpaper, 10 pt, conference]{ieeeconf}  

\IEEEoverridecommandlockouts                              

\usepackage{graphicx}
\usepackage{wrapfig}
\usepackage{amsmath}
\usepackage{algorithm}
\usepackage{algpseudocode}
\usepackage[percent]{overpic}
\usepackage{multirow}
\usepackage{booktabs}
\usepackage{cite}
\usepackage[T1]{fontenc}
\usepackage[font=small,labelfont=bf,tableposition=top]{caption}
\usepackage[usenames,dvipsnames]{xcolor}

\usepackage{comment}
\usepackage{url}
\usepackage{hyperref}
\hypersetup{
    colorlinks=true,
    linkcolor=blue,
    filecolor=magenta,      
    urlcolor=cyan,
    pdftitle={Real2Sim2Real Transfer for Control of Cable-driven Robots via a Differentiable Physics Engine},
    pdfpagemode=FullScreen,
    }

\DeclareCaptionLabelFormat{andtable}{#1~#2  \&  \tablename~\thetable}

\newenvironment{myitem}{\begin{list}{$\bullet$}
{\setlength{\topsep}{-5pt}
\setlength{\leftmargin}{6pt}
\setlength{\parsep}{0pt}
\setlength{\itemsep}{0pt}
\setlength{\partopsep}{0pt}}}%
{\end{list}}

\title{\LARGE \bf
Real2Sim2Real Transfer for Control of Cable-driven Robots\\ via a Differentiable Physics Engine
}

\author{Kun Wang$^{1}$, William R. Johnson III$^{2}$, Shiyang Lu$^{1}$, Xiaonan Huang$^{2}$, \\ Joran Booth$^{2}$, Rebecca Kramer-Bottiglio$^{2}$, Mridul Aanjaneya$^{1}$, and Kostas Bekris$^{1}$
\thanks{This work has been supported by NSF Robust Intelligence award 1956027. The opinions expressed here are those of the authors and do not reflect the positions of the sponsor. $^{1}$Computer Science, Rutgers University, NJ 08901, USA. Email: {\tt\small {\{kun.wang2012, shiyang.lu, mridul.aanjaneya, kostas.bekris\}@rutgers.edu}}. $^{2}$Mech. Eng. \& Material Sc., Yale University, New Haven, CT, USA.  Email: {\tt\small {\{will.johnson, xiaonan.huang, joran.booth, rebecca.kramer\}@yale.edu}.}}%
}

\begin{document}

\maketitle
\thispagestyle{empty}
\pagestyle{empty}

\vspace{-5mm}
\begin{abstract}

Tensegrity robots, composed of rigid rods and flexible cables, exhibit high strength-to-weight ratios and significant deformations, which enable them to navigate unstructured terrains and survive harsh impacts. They are hard to control, however, due to high dimensionality, complex dynamics, and a coupled  architecture. Physics-based simulation is a promising avenue for developing locomotion policies that can be transferred to real robots.  Nevertheless, modeling tensegrity robots is a complex task due to a substantial sim2real gap. To address this issue, this paper describes a Real2Sim2Real ({\tt R2S2R}) strategy for tensegrity robots. This strategy is based on a differentiable physics engine that can be trained given limited data from a real robot. These data include offline measurements of physical properties, such as mass and geometry for various robot components, and the observation of a trajectory using a random control policy.  With the data from the real robot, the engine can be iteratively refined
and used to discover locomotion policies that are directly transferable to the real robot. Beyond the {\tt R2S2R} pipeline, key contributions of this work include computing non-zero gradients at contact points, a loss function for matching tensegrity locomotion gaits, and a trajectory segmentation technique that avoids conflicts in gradient evaluation during training. Multiple iterations of the {\tt R2S2R} process are demonstrated and evaluated on a real 3-bar tensegrity robot.
\end{abstract}


\section{Introduction}

Tensegrity robots are actuated systems composed of rigid struts (rods) and flexible elements (cables) connected to form lightweight, deformable structures. Their natural compliance makes them adaptable and safe robots that are well-suited for many applications, such as manipulation~\cite{lessard2016bio}, locomotion~\cite{sabelhaus2018design}, morphing airfoils~\cite{chen2020design}, and spacecraft landing~\cite{bruce2014superball}. 

At the same time, tensegrity robots are difficult to accurately model and control due to their many degrees of freedom (DoF) and complex dynamics~\cite{shah2022tensegrity}. The difficulty in modeling has led some researchers to propose model-free solutions for learning control~\cite{Surovik2021AdaptiveTL}, but these strategies still encounter challenges since they require a large amount of training data, and collecting trajectories from tensegrities is time-consuming, cumbersome, and expensive. The authors have previously introduced a differentiable, but modular and explainable, physics engine~\cite{wang2020end, pmlr-v120-wang20b, wang2022recurrent} as a data-efficient tool to identify physical parameters and develop locomotion policies for tensegrity robots. Yet, the transfer of simulated policies to real hardware is typically impeded by the so-called simulation-to-reality (sim2real) gap~\cite{bousmalis2017closing, martinez2020unrealrox}. Sim2real transfer can be improved by tuning the simulation to minimize differences between predicted and real robot trajectories for the same controls.




To overcome the sim2real gap, many methods are applied~\cite{martinez2020unrealrox,murthy2020gradsim,collins2019quantifying, ramos2019rss, muratore2022neural}, however, the robot is assumed to be rigid~\cite{zeng2020tossingbot} as a prior. Systems that include compliance, such as tensegrity robots, are less frequently targeted, as compliance makes closing the sim2real gap more challenging. Only recently have Real2Sim2Real ({\tt R2S2R}) frameworks been developed that focus on deformable elements (e.g., a system that manipulates deformable cables~\cite{lim2021planar}). 


\begin{figure}[t]
    \centering
    \includegraphics[width=0.8\columnwidth]{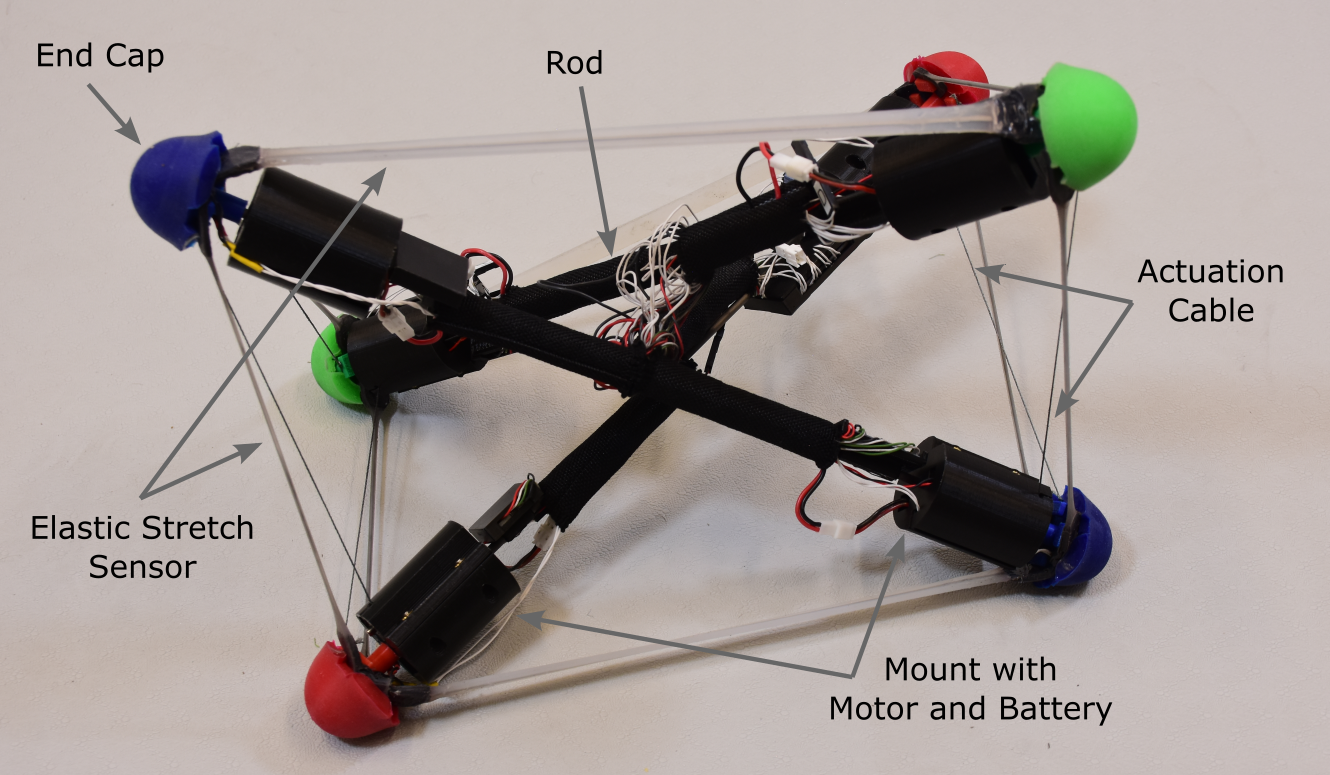}
    \vspace{-.05in}
    \caption{The target platform: a 3-bar prismatic tensegrity robot with 9 stretch sensors. The 6 short cables are contracted and extended by the motors in the 3D-printed housings on each rod.}
    \label{fig:labeled_tensegrity}
    \vspace{-.3in}
\end{figure}

This work applies the principle of {\tt R2S2R} to the control of deformable robots composed of both rigid and soft elements. More specifically, this work introduces an {\tt R2S2R} pipeline for tensegrity robots, through which a policy learned on a differentiable engine is transferred to a 3-bar tensegrity robot (Fig.~\ref{fig:labeled_tensegrity}) by first training the engine with data from the real robot. This paper contributes:





\begin{myitem}
    \item A complete pipeline for identifying the parameters of a differentiable simulator from real tensegrity robot trajectories, generating locomotion policies in the simulator, and transferring the policies back to the real robot.
    \item A method to compute non-zero gradients at contact points to enable efficient learning of contact parameters in the optimization step of the engine's identification process.
    \item A loss function and trajectory segmentation strategy to avoid conflicts in gradient direction during training under noisy observations. Non-convex trajectories lead to gradients with opposite directions at different time steps. Thus, the trajectory is segmented into convex segments to ensure the gradient directions computed are aligned.
\end{myitem}






\section{Related Work}
\label{sec:citations}
\vspace{-1mm}

Sim2real transfer has been applied in autonomous underwater vehicles~\cite{sethuramantowards}, drones~\cite{navardi2022toward,loquercio2019deep}, muscles~\cite{senagap}, quadruped robots~\cite{jabbourclosing}, soft robots~\cite{huang2020dynamic,huang2021numerical,goldberg2019planar}, and grasping manipulators~\cite{wen2022catgrasp}. To the best of the authors' knowledge, this is the first work that mitigates the sim2real gap for a tensegrity robot.


\textcolor{black}{Differentiable physics has been actively applied to system identification. Compared to artificial evolution approaches (e.g., genetic algorithms, particle swarm optimizations, and covariance matrix adaptation evolution strategies) and domain-randomization methods, gradient-based methods like differential physics are data-efficient and can lead to faster convergence for complex robot systems~\cite{de2018end,degrave2019differentiable}. 
We previously developed a differentiable engine for tensegrity robots~\cite{wang2020end, pmlr-v120-wang20b,wang2022recurrent}, although prior work has been limited to only sim2sim transfer and never demonstrated on a real tensegrity robot.}

Prior work on tensegrity locomotion~\cite{zhang2017deep,luo2018tensegrity,surovik2019adaptive} has achieved complex behaviors, sometimes on uneven terrain, using the NASA Tensegrity Robotics Toolkit (NTRT) simulator~\cite{NTRTSim}, which was manually tuned to match a real platform~\cite{mirletz2015towards,caluwaerts2014design}. Many prior approaches use reinforcement learning (RL) to learn policies given sparse inputs, which can be provided by onboard sensors~\cite{luo2018tensegrity} and aim to address the large data requirements of RL~\cite{surovik2019adaptive}, including by training in simulation. Simulated locomotion, however, is hard to replicate on a real platform, even after hand-tuning, which emphasizes the importance of training a simulator that can produce policies that can be transferred to a real system. A website accompanying this paper with videos and additional evaluation is available\footnotemark{}.

\footnotetext{An appendix with addtional material and accompanying multimedia can be found at: \href{https://sites.google.com/view/sim2real}{https://sites.google.com/view/sim2real}}

 
\vspace{-1mm}
\section{Robot Design}
\vspace{-1mm}
This work demonstrates and evaluates the {\tt R2S2R} pipeline on the untethered 3-bar prismatic tensegrity robot shown in Fig.~\ref{fig:labeled_tensegrity}. The tensegrity robot has a rod length of 36~cm, and it is driven by motors that extend and contract its cables to shift its center of mass.  The six short cables (three on each side) are actuated by the motors while the three longer tendons in the middle are passive elastic elements. These passive tendons double as stretch sensors, and there are also six sensor tendons in parallel with the six actuated cables. The design and characterization of the stretch sensors are detailed in previous work~\cite{johnson2022sensor}.
Each sensor is calibrated individually by fitting a linear model to map capacitance measurements to corresponding lengths~\cite{johnson2021integrated}. The stretch sensors are used both for feedback control and for pose estimation, as described in  section~\ref{sec:method_data_collect}. 



\vspace{-1mm}
\section{Real2Sim2Real Pipeline}
\vspace{-1mm}

\begin{figure}[t]
    \centering
    \includegraphics[width=1.0\columnwidth]{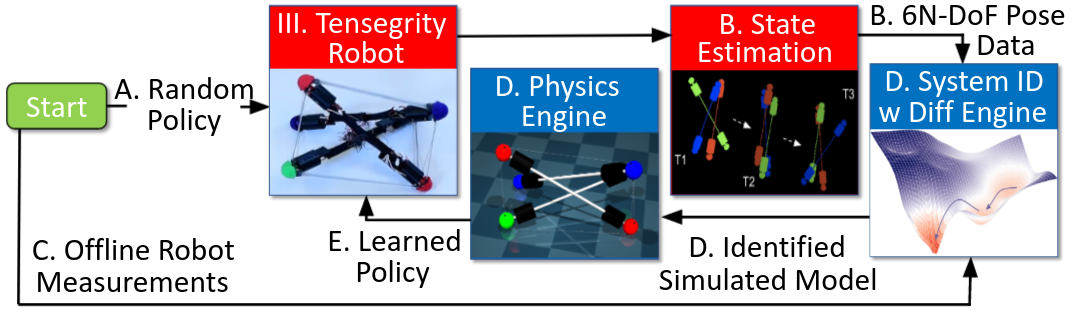}
    \vspace{-.2in}
    \caption{\textbf{R2S2R pipeline.} A randomly generated policy is executed first on the robot. An overhead RGB-D camera captures the trajectory to estimate the robot's state at each frame. Internal robot parameters are then identified with a differentiable physics engine given robot states and offline measurements. Given the identified model, new policies are generated and executed on the robot. The process can be repeated to further close the sim2real gap. Red blocks correspond to real-robot tasks; blue blocks correspond to simulation tasks. Labels correspond to sections in the text.}
    \label{fig:r2s2r_pipepine}
    \vspace{-.3in}
\end{figure}

The following discussion describes the various components of the proposed {\tt R2S2R} pipeline as outlined in Fig.~\ref{fig:r2s2r_pipepine}. 


\vspace{-1mm}
\subsection{Random Policy and Real World Execution}
\vspace{-1mm}


The first step in the {\tt R2S2R} pipeline is to generate a random policy and execute it on the real robot. The policy is defined as a list of robot cable lengths (0 means fully contracted and 1 means fully extended) that represent a series of target robot shapes. To transition between these high-level shapes, a PID controller extends or contracts the actuated cables by sending low-level control signals based on the feedback from the robot cable length sensors until the robot reaches the next target shape. All policies start from the robot rest state where the robot sits on the ground with the six actuated cables fully extended, as shown in Fig. \ref{fig:labeled_tensegrity}. 

\vspace{-1mm}
\subsection{State Estimation} \label{sec:method_data_collect}
\vspace{-1mm}

The next step is to estimate the robot states, i.e., the position and orientation of each rod, at each time frame. We adopt a state estimation method~\cite{lu20226ndof} designed for tensegrity robots. This method tracks the 6-DoF pose of each rod given a multi-modal input sequence of RGB images, depth images, and measured cable lengths from the onboard stretch sensors. The tracking method incorporates a variety of physical constraints, and it performs state estimation with high accuracy and robustness to self-occlusions. Further,
RANSAC~\cite{fischler1981random} is used for ground detection so that contacts between end caps and the ground could be found(Section \ref{sec:method_sys_id}).

\begin{figure}[ht]
    \centering
    \vspace{-.15in}
    \includegraphics[width=1\columnwidth]{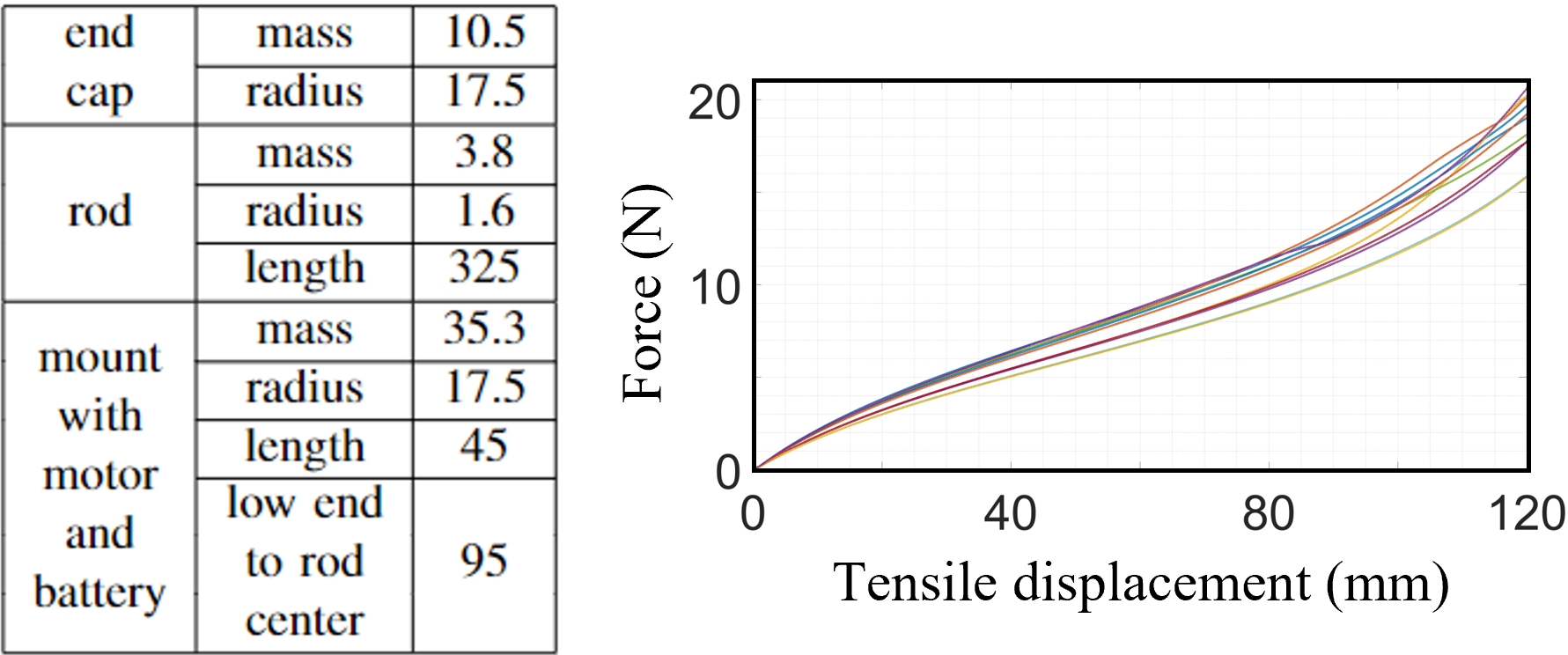}
    \vspace{-.2in}
    \caption{(Left) Measured physical parameters. Mass and length are in grams and mm, respectively. (Right) Force-displacement curves for long tendons (N = 11). A third-order polynomial was fit to each curve, and the average was used in the simulator.}
    \label{fig:robot_measurement}
    \vspace{-.2in}
\end{figure}

\vspace{-1mm}
\subsection{Offline Robot Measurements} \label{sec:method_robot_measure}
\vspace{-1mm}

Physical parameters inside a simulator (e.g., mass, radius of the end caps, lengths of the rods, inertia tensor, stiffness coefficient, coefficient of friction, etc.) are key to modeling a real robot with high fidelity. However, identifying all of these parameters in one optimization is challenging.
Instead, we measure some of these parameters offline and use them as inputs to the simulator.  The measured mass and physical dimensions of the robot's components are listed in Fig.~\ref{fig:robot_measurement}.
We also collected stiffness data for the long, passive tendons. A batch of 11 sensor tendons was measured on a materials testing system (Instron 3345) and fit with a third-order polynomial; we averaged the curves to input a single model into the simulator. The short, actuated cables are inelastic and modeled with high stiffness in the simulation.

\vspace{-1mm}
\subsection{System Identification}\label{sec:method_sys_id}
\vspace{-1mm}


Physical parameters like the coefficient of friction and the speed of the robot's motors are difficult to measure offline.
We have previously introduced a differentiable physics engine that uses gradient descent to identify system parameters from ground truth data~\cite{wang2020end, pmlr-v120-wang20b, wang2022recurrent}.  However, this engine has been demonstrated only in simulation.  This section describes the key differences that are necessary for system identification when a real robot is the target platform.

\begin{figure}[!htpb]
    \vspace{-3mm}
    \centering
    \includegraphics[width=1\columnwidth]{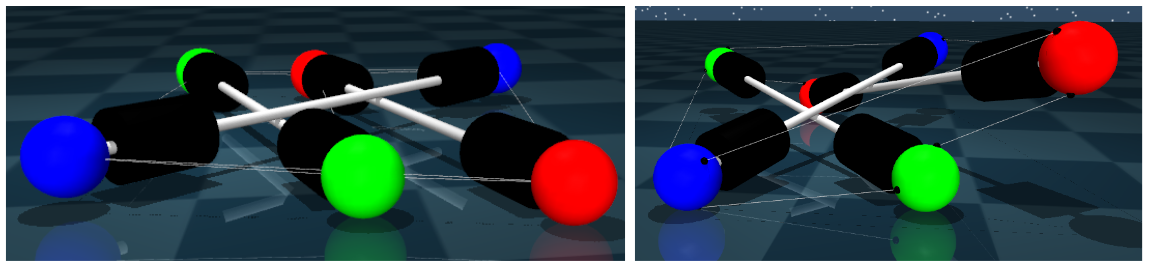}
    \vspace{-6mm}
    \caption{(Left) After certain controls, the robot collapses onto the ground if all cables are attached to the end of the rods. (Right) With the same control sequences, the robot does not collapse if the cables are attached on the surface of the end caps.}
    \label{fig:diff_cable_atttachment}
    \vspace{-3mm}
\end{figure}

Some parameters, such as the cable attachment points, are not measured on the real robot but are crucial for the success of the simulation. The benchmark tensegrity simulator, NTRT~\cite{NTRTSim},
attaches cables at the ends of each rod, which simplifies the modeling process. However, this attachment strategy can lead the simulated robot to collapse 
and get stuck, unable to
recover because all the actuated cables are coplanar, as shown in Figure~\ref{fig:diff_cable_atttachment} (Left). Although, this never happens to the real robot.
Instead, we put the cable attachment points on the surface of the end caps. This setup prevents the collapsing and results in a more stable simulation, as shown in Figure~\ref{fig:diff_cable_atttachment} (Right).
Note that these attachment points are not available in the observations; we implement a heuristic to infer their positions.

\begin{figure}[!htpb]
    \centering
    \includegraphics[height=0.3\columnwidth]{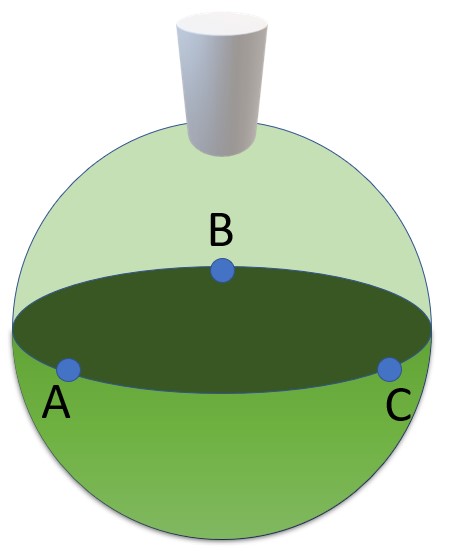}
    \includegraphics[height=0.3\columnwidth]{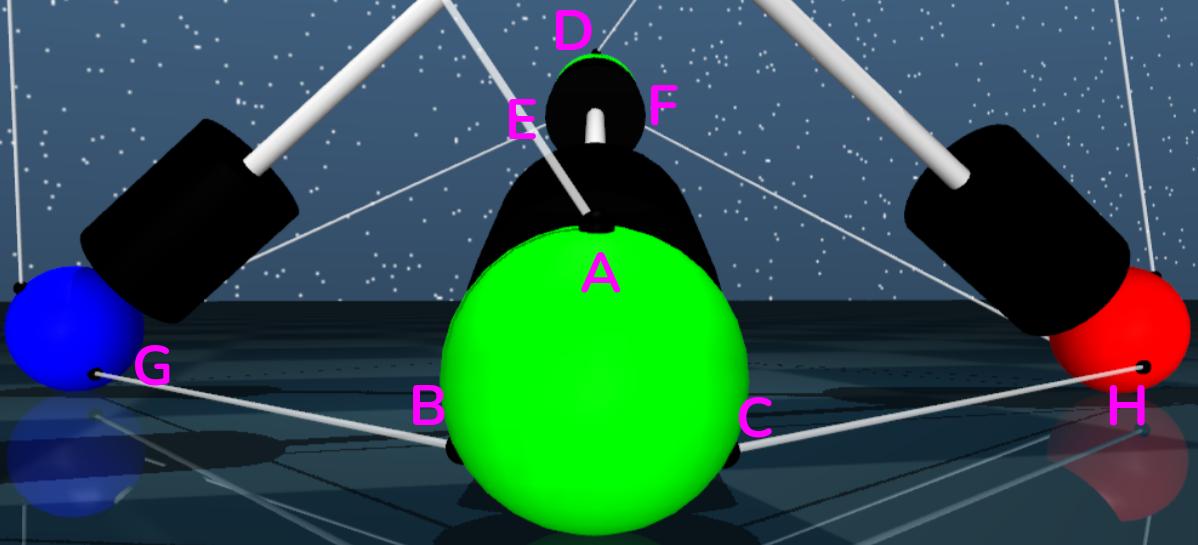}
    \vspace{-1mm}
    \caption{Cable attachment points on the end caps in the rest state.}
    \vspace{-5mm}
    \label{fig:cable_att_p}
\end{figure}

Each rod has six attachment points that are distributed symmetrically and evenly on two end caps. Attachment points are placed evenly at robot construction every $2\pi/3$ along the disk that is perpendicular to the rod. So it is possible to infer all attachment points from only one of them. At the robot rest state, one rod is in the center and the other two rods are on the side. Referring to  Fig.~\ref{fig:cable_att_p}, we first compute point $A$ directly, which is on top of the end cap. Then $B,C$, can be inferred from it. Since the segments $AD, BE, CF$ are parallel to the rod, points $D,E,F$ on the other end are computed based on points $A,B,C$. Points $G,H$ are closest to points $B,C$. 

Before each trajectory, the real robot resets itself to the rest state. We also reset the simulator to align with the real robot's initial state 
using the data collected at time $t=0$, including both cable lengths and end cap positions.
The initial pose generated by the state estimation algorithm (Section~\ref{sec:method_data_collect}) may not have the robot perfectly resting on the ground due to the limitations of the sensors.  
Thus, we added gravity to the simulation to force the lowest end caps to come in contact with the ground.


After initializing the robot in simulation, we use the differentiable physics engine to identify the remaining physical parameters by performing gradient descent over a trajectory.  To accomplish this task, we introduce a detachment method to compute nonzero gradients at contact points in order to efficiently learn contact parameters. Furthermore, the long-horizon trajectories and noisy observations from the real robot can lead to conflicts in the gradient direction.  To combat these conflicts, we introduce the Key Frame Loss (KFL) function to segment the trajectories and compute losses at gait transitions. The detachment method and KFL are described in Sections~\ref{sec:detach_contact_grads} and~\ref{sec:key_frame_loss}, respectively. 

\vspace{-1mm}
\subsection{Developing Policies with Symmetry Reduction}
\vspace{-1mm}

Policy search on tensegrity robots is challenging due to combinatorial explosion since a sequence of control signals for each actuation cable with unknown time horizon are need to be discovered. 
Symmetry reduction has been developed for 6-bar tensegrity robots~\cite{surovik2019adaptive,surovid2018any} to reduce the policy search space.
Here, we apply symmetry reduction to our 3-bar tensegrity robot and combine it with hybrid control, where a high-level planner outputs the next target state and a low-level PID controller commands the cable lengths. 


\begin{figure}[!htpb]
    \vspace{-3mm}
    \centering
    \includegraphics[height=0.3\columnwidth]{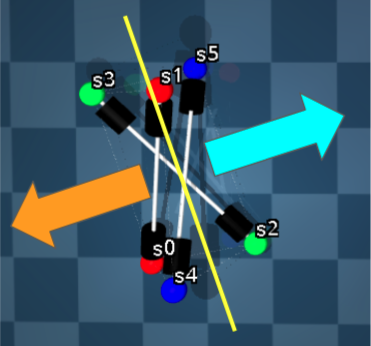}
    \includegraphics[height=0.28\columnwidth]{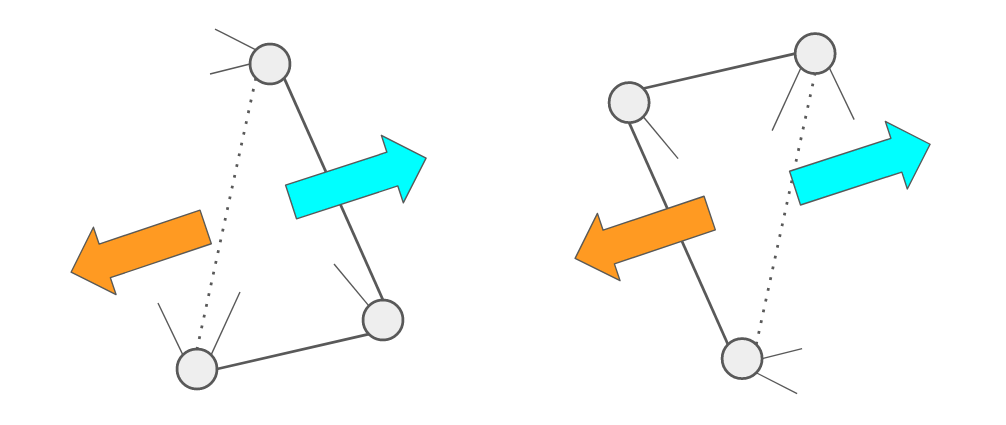}
    \vspace{-5mm}
    \caption{(Left) The robot principle axis (yellow) is the average direction of the three rods. The forward direction is indicated by the cyan arrow; the backward direction is indicated by the orange arrow. (Right) Two robot poses, with different types of bottom triangles, are mapped after rotating $\pi$ radians around the ground normal.}
    \label{fig:symmetry_reduction_control}
    \vspace{-3mm}
\end{figure}

To reduce the state space via symmetry reduction, we use a relabeling scheme defined with respect to the support triangle. We define the robot's state in Fig.~\ref{fig:symmetry_reduction_control} (Left) as the reference state, from which we can rotate the robot about its principle axis (the yellow line in Fig.~\ref{fig:symmetry_reduction_control}) by intervals of $\pi/6$ to get 2 types of support triangles with opposite forward and backward directions. To map from one type of support triangle to the other, we rotate the robot by $\pi$ around the ground normal. After all rotations, we identify the mapping by computing the closest end caps to the reference state.


To generate gaits, i.e., time-ordered sequences of target poses, we conduct a graph search starting from the rest pose. These short gaits are mapped to long control policies via symmetry expansion. The nodes of the graph are robot poses represented by binary cable lengths (0 is retracted and 1 is extended), and the edges of the graph are cable controls (retract, extend, or hold).  To limit the search space, we search to a depth of 4. The root node and leaf nodes are in the same rest pose where all cables are fully extended.  We use the center of mass displacement as the reward function to generate one forward and one backward rolling gait. We use principal axis rotation as the reward function to generate clockwise and counterclockwise turning gaits. 



\vspace{-1mm}
\subsection{Detachment Approach for Contact Gradients}
\vspace{-1mm}
\label{sec:detach_contact_grads}

Impulse-based dynamics can't be directly applied to differentiable simulations due to a zero gradient issue~\cite{werling2021fast}. We propose a ``detachment'' method to obtain nonzero gradients for contact parameters $e, \mu$. Passive forces, such as restitution and friction generate equal gradients in opposite directions, and gradients of these forces should not be backpropagated. The detach() function~\cite{detach} returns a new tensor detached from the computational graph, truncating the gradient propagation path. 


Our detachment method can be applied to learn the restitution $e$ and the force $F = mg - N$, which is shown in Algorithm~\ref{alg:detach_restitution}. The difference of speed ($\Delta v$) and position ($\Delta x$) at time ${t+1}$ and time $t$ are detached to stop the passive force gradient from backpropagating. Otherwise, the gradient would be $d x_{t+1}/ d e = 0.$

\vspace{-3mm}
\begin{algorithm}
\caption{Detachment Method for Restitution}\label{alg:detach_restitution}
\begin{algorithmic}
\State $v_{t+1}' = v_t - g \Delta t + F/m\Delta t$
\State $\Delta v = -(1+e)v_{t+1}'.detach()$
\State $v_{t+1} = v_{t+1}' + \Delta v$
\If { $x_t < ground$}
    \State {$\Delta x =  ground-x_t$}
    \State {$x_{t+1} = x_t + v_{t+1} \Delta t + \Delta x.detach()$}
\Else
    \State {$x_{t+1} = x_t + v_{t+1} \Delta t$}
\EndIf
\end{algorithmic}
\end{algorithm}
\vspace{-3mm}


The detachment approach is also used for learning friction parameters, as shown in Algorithm~\ref{alg:detach_mu}. We detach $\Delta v$ because the friction force is a passive force. We add the term $-\Delta p_f  + \Delta p_f.detach()$ to pass the gradient to $\mu$ in the case where there is only static friction. \textcolor{black}{This term guarantees that $\mu$ can always be updated by the gradient, no matter if the initial $\mu$ is higher or lower than the actual one. The evaluation of the contact gradients and integration with Time of Impact (ToI) computations can be found in the accompanying appendix available online\footnotemark[\value{footnote}]. }

\begin{algorithm}
\caption{Detachment Method in Friction Computation}\label{alg:detach_mu}
\begin{algorithmic}
\State $v_{t+1}' = v_t + F/m \Delta t$
\State $\Delta v = -v_{t+1}'.detach()$
\State $\Delta p_f = \mu N \Delta t / m$
\If { $|\Delta v| > \Delta p_f $}
    $\Delta v =  \Delta v * \Delta p_f  /  |\Delta v| $
\Else
    \ $\Delta v =  \Delta v - \Delta p_f  + \Delta p_f.detach()$
\EndIf
\State {$v_{t+1} = v_t + F/m \Delta t + \Delta v$}
\State {$x_{t+1} = x_t + v_{t+1} \Delta t$}
\end{algorithmic}
\end{algorithm}
\vspace{-0mm}


\vspace{-0mm}
\subsection{Key Frame Loss}
\label{sec:key_frame_loss}
\vspace{-1mm}

System identification with long trajectories is difficult due to sensing noise and the non-monotonicity of trajectories. 

\begin{figure}[!htpb]
    \vspace{-2mm}
    \centering
    \includegraphics[width=\columnwidth]{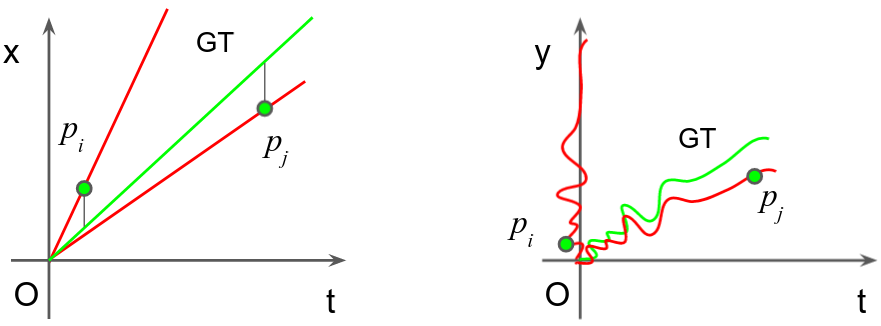}
    \vspace{-6mm}
    \caption{Fitting the ground truth line (green) with two noisy observations, $p_i$ and $p_j$. Although $p_i$ and $p_j$ have the same sensing error, the fitted line (red) by $p_j$ is better than that by $p_i$. $x=vt$ is a linear function (Left). $y=f(t)$ is a non-convex function (Right).}
    \label{fig:noisy_obs_error}
    \vspace{-5mm}
\end{figure}

\begin{figure}[!htpb]
\vspace{-3mm}
    \centering
    \includegraphics[width=\columnwidth]{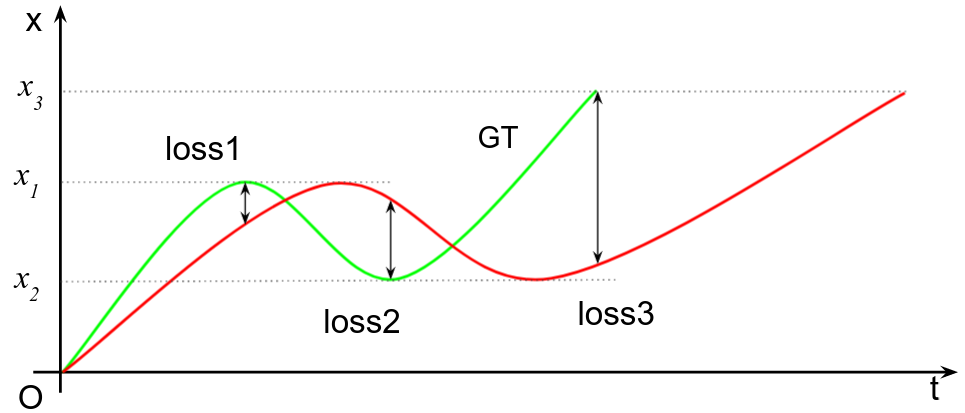}
    \vspace{-6mm}
    \caption{The ground truth trajectory (green) includes three targets, $x_1$, $x_2$ and $x_3$. The unidentified engine (red) has a lower speed and takes more time to reach the targets. Three losses are computed at these targets. The gradients at loss1 and loss3 pull up the red line; however, the gradient at loss2 drags it down.}
    \vspace{-3mm}
    \label{fig:contradicted_gradient}
\end{figure}

As shown in Fig.~\ref{fig:noisy_obs_error}, the later observations are preferable to reduce the fitting error. 
Consider a simple example of a 1D trajectory from $x=vt$ with length $T$. We are given two data points, $p_i = (t_i, x_i)$ and $p_j = (t_j, x_j)$. $t_i$ is close to $0$ and $t_j$ is close to $T$. $x_i$ and $x_j$ have the same observation error relative to the ground truth. $v$ is the parameter to estimate. Fig.~\ref{fig:noisy_obs_error} (Left) shows that the fitted line by $p_j$ is better than that by $p_i$. Moreover, if $x=f(t)$ is a non-\textcolor{black}{monotonic} function, which is closer to reality for tensegrity robot trajectories, the fitted line by $p_i$ is worse as shown in Fig.~\ref{fig:noisy_obs_error} (Right). The trajectory of an end cap is noisier around time $t=0$ because of the sudden motor activation. Note that the loss landscape may be non-convex due to contradictions in gradient directions, which can interfere with the optimization step. As shown in Fig.~\ref{fig:contradicted_gradient}, three losses yield opposite gradient directions.

The multiple-shooting (MS) method has been applied for system identification with long trajectories~\cite{heiden2022pds}. MS, however, needs full robot observations (i.e., positions and velocities of each end cap) to initialize the simulator. The defects may lead to contradicting gradient directions.

\begin{figure*}[t]
    \centering
    \includegraphics[width=2.0\columnwidth]{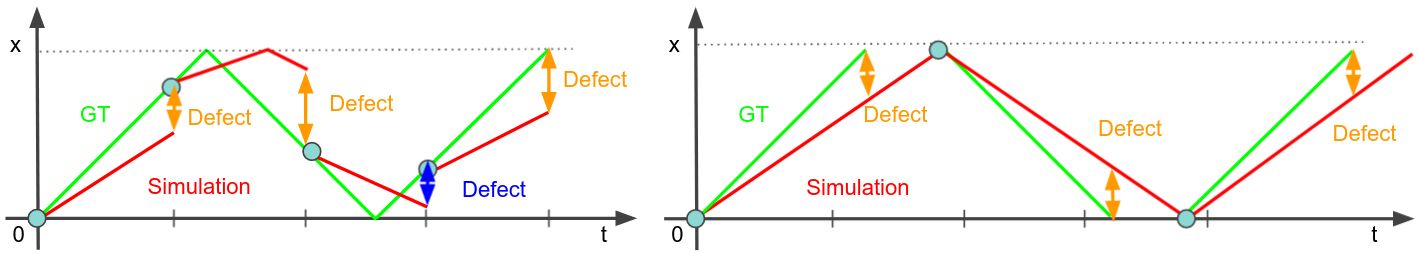}
    \vspace{-.05in}
    \caption{(Left) The multiple-shooting (MS) method needs both positions and velocities at each time step so the simulator can align with the ground truth at intermediate time steps. It uses the entire length of a time window to compute defects between the ground truth and the simulator. The window size is empirical, and the defects may provide contradicting gradient directions. (Right) The KFL method only needs positions at each time step. The trajectory is split into segments based on unidirectional control intervals. The time step at the end of the control signal is the key frame where the defects are computed. The window size is the same size as the control interval, and the defects provide consistent gradient directions.}
    \label{fig:ms_vs_fkl}
    \vspace{-.2in}
\end{figure*}
We introduce the Key Frame Loss (KFL) function, which splits the robot trajectory into a list of monotonic segments using unidirectional control intervals. We only consider the loss from the last frame in each segment. There are four departures of KFL from MS: 1) KFL \textcolor{black}{only needs partial observation}, i.e., it only requires end cap positions, not their velocities; 2) In the intermediate time steps, we do not aim to recover the simulator from the partially observed state since we do not know velocities; 3) The observation frequency does not have to be constant; 4) The shooting time window size is not fixed. Fig.~\ref{fig:ms_vs_fkl} provides a comparison between MS and KFL.

\textcolor{black}{The benefits of KFL are that it 1) eliminates the factor of time, 2) makes system identification robust to observation noise, and 3) ensures the correct gradient direction. This strategy can be generalized to other optimization problems using trajectories as ground truth data: trajectories can be split into monotonic segments, and the KFL can be applied at the final frame of each segment like in this work.}

\begin{figure}[!htpb]
    \centering
    \includegraphics[width=\columnwidth]{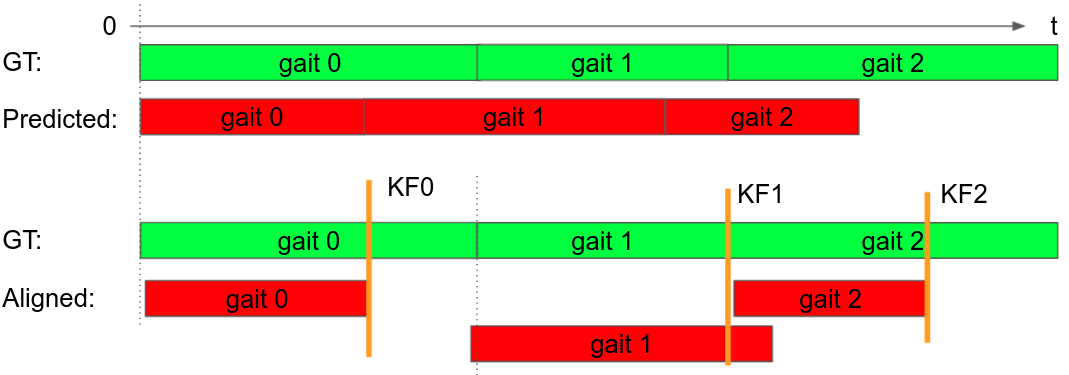}
    \vspace{-.2in}
    \caption{(Above) The ground truth (GT) and engine-predicted trajectory are split into three gaits based on the control intervals. (Below) The gaits are aligned at the starting points and their key frames (KF0, KF1, KF2) are the last time frames of each intersection. The predicted trajectory takes less time in gait 0 and 2, but more time in 1, which leads to contradicting gradient directions.}
    \label{fig:key_frame_loss}
    \vspace{-4mm}
\end{figure}

To get the key frames, we split the trajectory into gaits for \textcolor{black}{monotonic} segmentation, as  in Fig.~\ref{fig:key_frame_loss}. In each gait, each cable follows only one of the possible controls: retracting, holding, or extending. These controls form a convex trajectory segment, avoiding the zig-zag trajectory that leads to contradicting gradients shown in Fig.~\ref{fig:contradicted_gradient}.
The key frame (KF) is the last time step of each gait. Consider an example trajectory with three gaits, as shown in Fig.~\ref{fig:key_frame_loss}. Both the ground truth (red) and the engine-predicted trajectory are split into three gaits. The corresponding gaits are remapped, and the KFs are the last time frames of their intersections. The KFL is the system difference, i.e., the sum of the differences in the positions of the end caps at the KF.
We select the last time step because 1) the observation noise around the start of each gait may lead to worse fitting as shown in Fig.~\ref{fig:noisy_obs_error} and 2) the gait transition point is where one gait is finished; however, whether the gait execution is complete is determined by measurements from the cable length sensors, so it is possible the gait transition occurs earlier or later than predicted, as shown by the gaps in Fig.~\ref{fig:key_frame_loss}.

Losses from different gaits may have contradicting gradients. In Fig.~\ref{fig:key_frame_loss}, the predicted trajectory takes less time for gait 0 and 2, but it takes more time for gait 1. These contradicting gradients can confuse the optimizer, as it will not know whether to speed up or slow down. Some possible reasons for these time differences are 1) The ground truth timestamp is noisy due to the limitations of the sensors; 2) The noisy cable length sensor readings lead to early or later gait transitions; 3) The starting states are not perfectly aligned; 4) There are gaps between the simplified physics engine and the real robot. To solve the problem, we adopt a mask filter to handle the contradicting gradients. 
Our heuristic states that the identified engine should execute the same gaits on the robot with similar time and behavior. Then, we take the execution time of the whole trajectory as an indicator to filter out the opposing gradients. For example, in Fig.~\ref{fig:key_frame_loss}, since the predicted trajectory takes a shorter length of time than the ground truth, we only consider key frame losses of gait 0 and gait 2 to slow down the engine, and we ignore the loss from gait 1. Additional evaluation about the KFL can be found in the link to the appendix\footnotemark[\value{footnote}]. 
\section{Experimental Results}
\label{sec:result}
Our experiments evaluate the {\tt R2S2R} pipeline both in simulation and on the real 3-bar tensegrity robot.  Section~\ref{sec:synthetic_trajectories} evaluates key components of the proposed process, i.e., the detachment strategy and the Key Frame Loss (KFL), using synthetic data. 
Section~\ref{sec:pipeline_evaluation} evaluates the full {\tt R2S2R} pipeline on the 3-bar tensegrity robot shown in Fig.~\ref{fig:labeled_tensegrity} with the dimensions in Fig.~\ref{fig:robot_measurement}. Section~\ref{sec:2nd_iteration} shows the additional improvement when a second iteration of the {\tt R2S2R} pipeline is performed. 

\subsection{System Identification with Synthetic Trajectories}
\label{sec:synthetic_trajectories}
\begin{figure}[ht]
    \centering
    \vspace{-.15in}
    \includegraphics[width=\columnwidth]{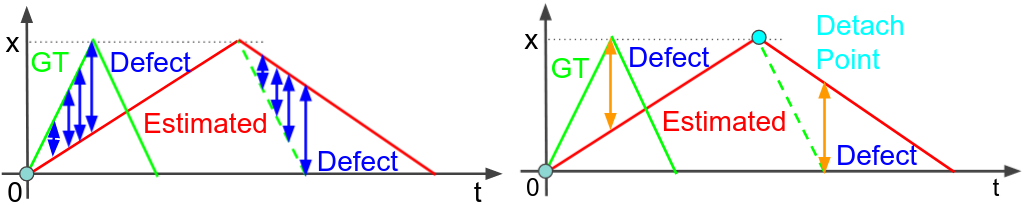}
    \vspace{-.25in}
    \caption{A trajectory starts from the origin, moves along $x$, and then returns to the origin.  This trajectory can be split into 2 windows. The ground truth trajectory (green) takes less time than the estimated one (red). The second window of the ground truth is shifted to the right to compute defects. (Left) The All Step loss function averages the defects of all time steps in each window. (Right) The Last Step loss function averages defects from the last time step in each window. KFL also detaches the system state at the transition point (cyan) of 2 windows~\cite{detach}.}
    \label{fig:ms_vs_our_loss_cmp}
    \vspace{-.15in}
\end{figure}

We evaluate KFL by comparing three different loss functions:  1) The average defects from all time steps (\emph{All Step}) in each window; 2) The average defects of the last time step (\emph{Last Step}) in each window; 3) Our method (Ours) with KFL, which averages defects of the last time step in each window and detaches the observations between each window. The difference between these three methods is shown in Fig.~\ref{fig:ms_vs_our_loss_cmp}.

\begin{figure}[ht]
    \vspace{-.1in}
    \centering
    \includegraphics[width=0.48\columnwidth]{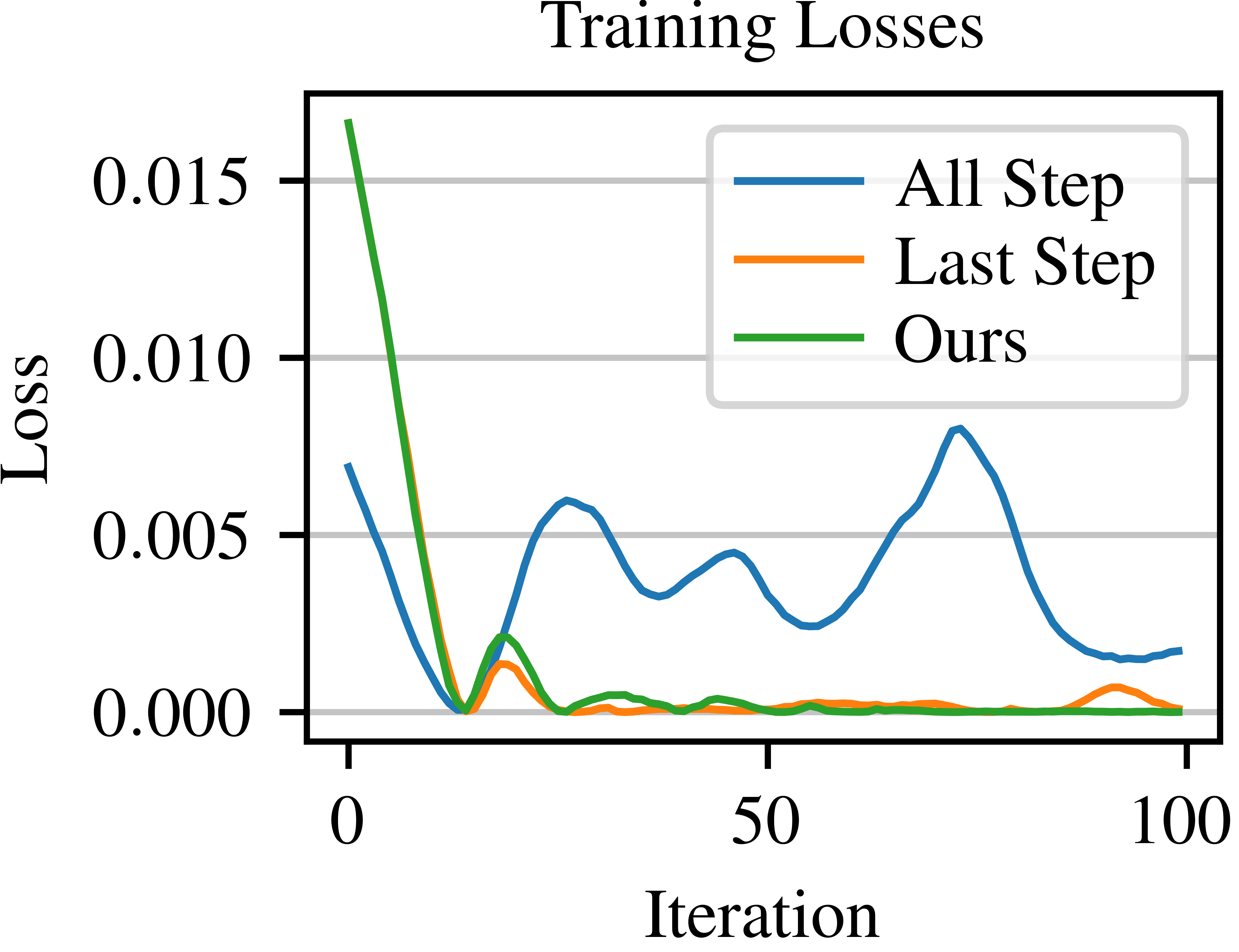}
    \includegraphics[width=0.48\columnwidth]{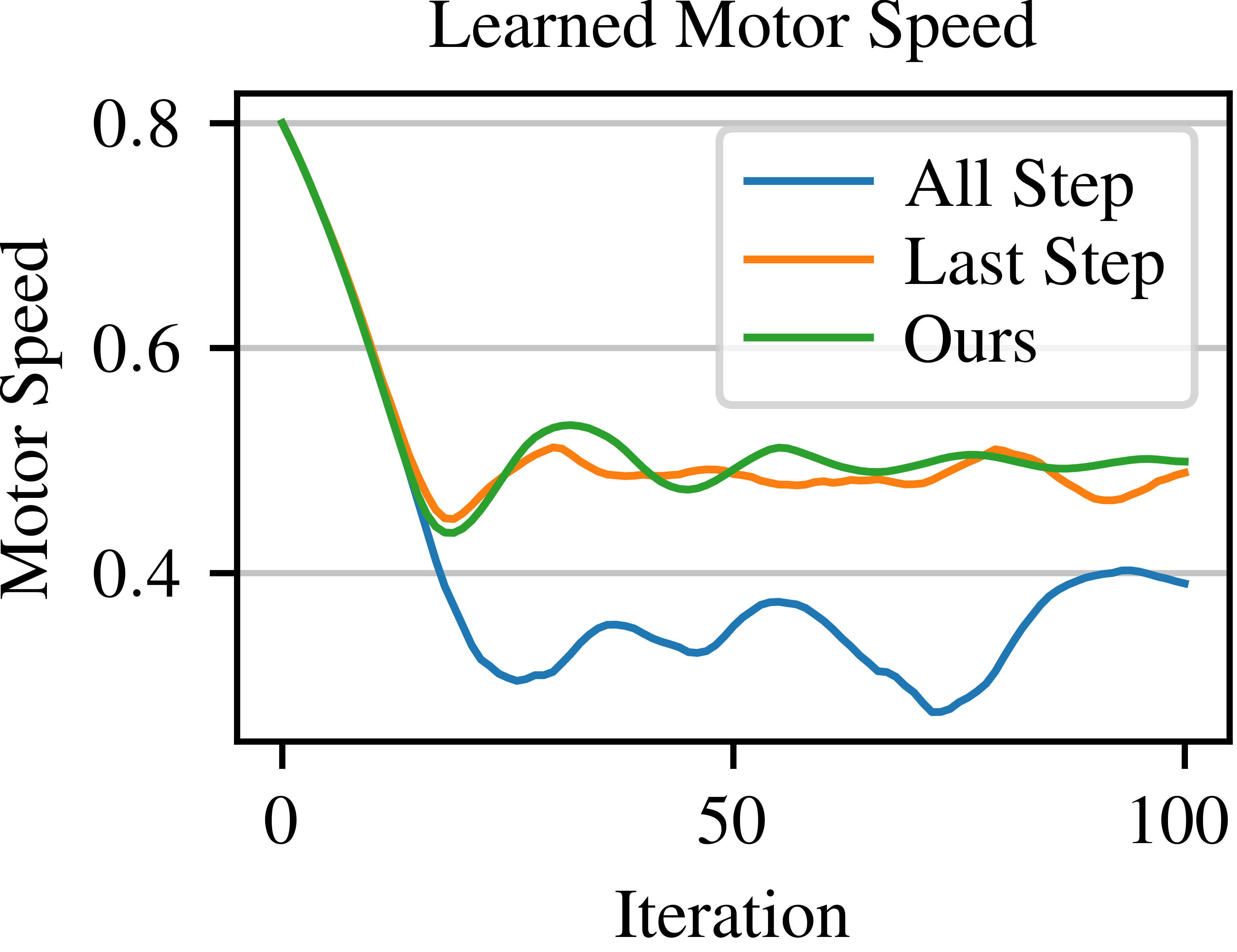}
    \vspace{-.05in}
    \caption{To identify the motor speed parameter, three loss functions are considered: All Step (blue), Last Step (orange), and KFL loss with state detachment (green). KFL loss converges better.}
    \label{fig:sim2sim_eval}
    \vspace{-.2in}
\end{figure}

\textcolor{black}{To simplify the problem, only motor speed is identified in this task (in addition to motor speed, the friction coefficient is also identified in V-B).} A synthetic trajectory where the motor speed is set to 0.5 is used as ground truth. The defects are the differences in end cap positions. An Adam optimizer with learning rate 0.1 is applied. Fig.~\ref{fig:sim2sim_eval} shows that our method converges fast, smooth and stable. The curve of the Last Step method fluctuates more due to the gradient propagation across windows, even if the motor speed is close to the target in later iterations. This shows the neeed for detaching. The multiple windows approach considering all time steps performs worse.

\subsection{\textcolor{black}{Complete Evaluation of the {\tt R2S2R} Pipeline}}
\label{sec:pipeline_evaluation}
\begin{figure}[ht]
    \centering
    \includegraphics[width=\columnwidth]{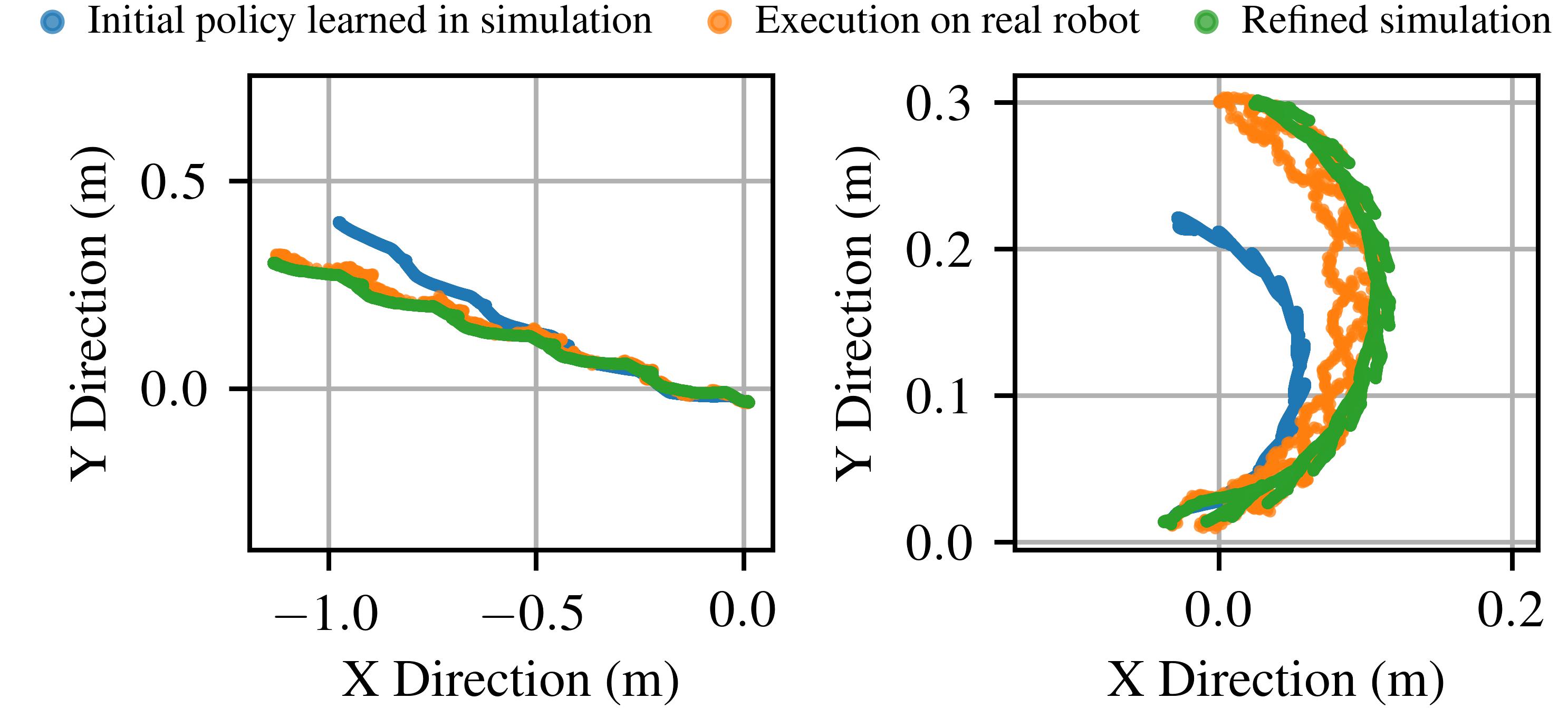}
    \caption{Trajectories in simulation and on the real robot for the same policy. The open-loop trajectories start at (0, 0). The robot trajectory is observed in order to generate new data to further refine the simulator. (Left) The output for a straight-line trajectory of the center of mass. The policy executes forward and backward motion in the symmetry-reduction frame. (Right) The output from a counterclockwise turning policy.}
    \label{fig:three_policy_real_vs_sim}       
    \vspace{-.3in}
\end{figure}

The {\tt R2S2R} pipeline has been applied twice to show continuous improvement each iteration. 
In the first iteration, a single robot trajectory given random controls is used to identify the physics engine parameters, including motor speed and friction coefficient. From the identified engine, we generate three policies corresponding to ``forward rolling,'' ``backward rolling,'' and ``counterclockwise turning'' behaviors. These policies are mapped to two long, open-loop gaits with symmetry reduction: a straight gait (Straight) composed of forward and backward rolling policies and a counterclockwise turning gait (CCW Turning) composed of ``counterclockwise turning'' polices. These two gaits are executed in simulation and on the real robot, and these executions are shown as \textcolor{blue}{blue} and \textcolor{orange}{orange} lines, respectively, in Figure~\ref{fig:three_policy_real_vs_sim}. \textcolor{black}{
We also execute these gaits starting from different positions on the ground to test the repeatability on the real platform (Table ~\ref{tbl:long_trajectory_repeatability}). The Center of Mass (CoM) and orientation are measured at the end of each trajectory.
We observe greater variability in the CCW Turning gait.  In the simulator, this gait capitalizes on uniform friction to rotate the robot's principal axis; however, in the real world, the friction between the ground and the end caps is nonuniform.  This discrepancy explains the higher variability.}

\begin{table}[!htpb]
\vspace{-.1in}
\caption{Real Robot Gait Execution Repeatability Test}
\vspace{-2mm}
\centering
\resizebox{\columnwidth}{!}{
\begin{tabular}{ c|c c c c} 
\toprule
 Metric & \multicolumn{2}{c}{End CoM Position (m)} & \multicolumn{2}{c}{End Orientation (radian)} \\
 \cmidrule{1-5}
Trajectory & Straight & CCW Turning & Straight & CCW Turning\\ 
 \midrule 
 Mean $\pm$ std & [-1.14$\pm$ 0.02, 0.29$\pm$0.05]& [0.01$\pm$0.02, 0.29$\pm$0.02] & 0.136$\pm$0.07  & 2.7$\pm$ 0.23\\
\bottomrule
\end{tabular}
}
\label{tbl:long_trajectory_repeatability}
\end{table}

\subsection{\textcolor{black}{Continuous Improvement with {\tt R2S2R}}}
\label{sec:2nd_iteration}

The second iteration re-identifies the engine using the two real robot trajectories from the first iteration, and then we execute these policies again in simulation to show how the {\tt R2S2R} pipeline reduces the sim2real gap. The trajectories sampled from the refined simulation are plotted as \textcolor{green}{green} lines in Fig.~\ref{fig:three_policy_real_vs_sim}. The robot position and orientation at the end of the trajectories are compared in Table~\ref{tbl:long_trajectory_real_vs_sim}. Generally, the CoM and orientation errors go down in the refined simulation; however, the slightly increased CCW Turning orientation error shows the limitations of simulating nonuniform environment friction with a uniform friction coefficient.

\begin{table}[!htpb]
\vspace{-.1in}
\caption{Iterative Refined Simulation with Real Data}
\vspace{-2mm}
\centering
\resizebox{\columnwidth}{!}{
\begin{tabular}{ c|c c c c} 
\toprule
Metric & \multicolumn{2}{c}{End CoM Position (m)} & \multicolumn{2}{c}{End Orientation (radian)} \\
 \cmidrule{1-5}
 Trajectory & Straight & CCW Turning & Straight & CCW Turning\\ 
 \midrule 
 Initial Simulation & [-0.97, 0.40]& [-0.02, 0.22] & 0.31 & 2.36\\ 
 Real Robot Execution & [-1.11, 0.32]& [0.02, 0.29] & 0.02 & 1.78\\ 
 Refined Simulation & [-1.11, 0.29]& [0.05, 0.30] & 0.12 & 2.45\\ 
 \midrule
 Initial Error & 0.16 & 0.26 & 0.09 & 0.04 \\
 Refined Error & 0.03 & 0.06 & 0.03 & 0.13 \\
\bottomrule
\end{tabular}
}
\label{tbl:long_trajectory_real_vs_sim}
\vspace{-.1in}
\end{table}

\begin{figure}[ht]
    \centering
    \includegraphics[width=\columnwidth]{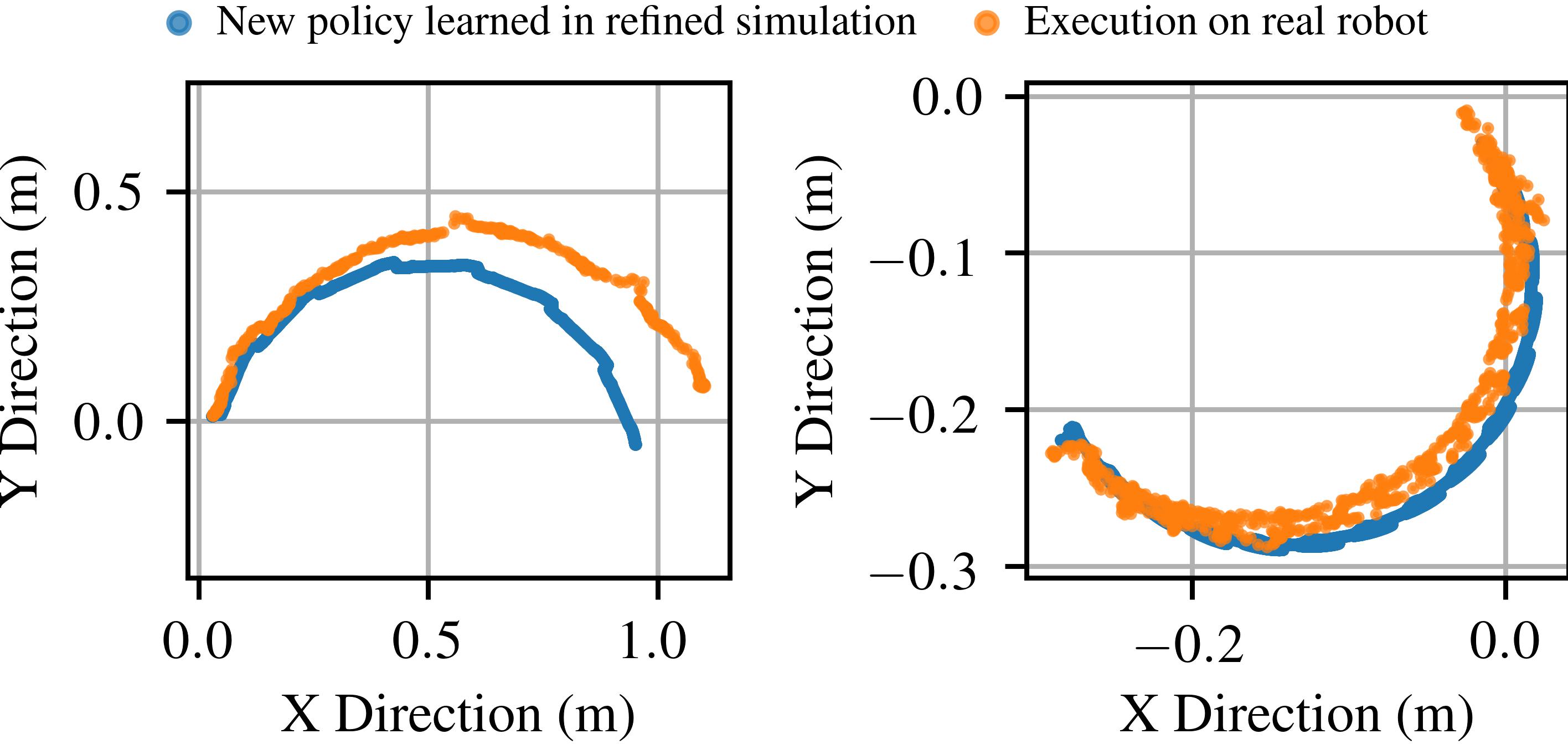}
    \vspace{-.2in}
    \caption{Two new clockwise turning trajectories are executed in simulation and on the real robot. The open-loop trajectories start at (0, 0). (Left) The arch turning trajectory. (Right) The in-place turning trajectory.}
    \label{fig:new_cw_policy}       
    \vspace{-.05in}
\end{figure}

\begin{figure}[ht]
    \centering
    \includegraphics[width=\columnwidth]{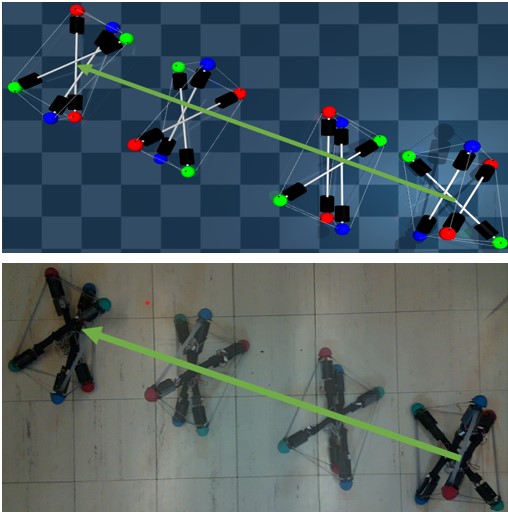}
    \vspace{-.2in}
    \caption{The frame-by-frame comparison of the straight trajectory execution in simulation and on the real robot.}
    \label{fig:straight_sim_vs_real}       
    \vspace{-.15in}
\end{figure}
\begin{figure}[ht]
    \centering
    \includegraphics[width=\columnwidth]{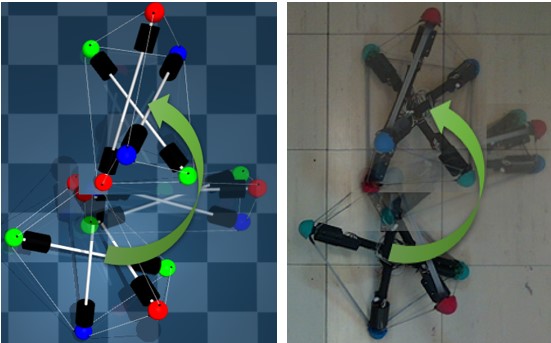}
    \vspace{-.2in}
    \caption{The frame-by-frame comparison of the counterclockwise turning trajectory execution in simulation and on the real robot.}
    \label{fig:ccw_sim_vs_real}       
    \vspace{-.05in}
\end{figure}

Qualitatively, the trajectory executed on the robot has the same behavior as the initial simulation. After the second iteration, the simulation trajectory is even more similar to the real robot trajectory, as shown in Fig~\ref{fig:straight_sim_vs_real} and \ref{fig:ccw_sim_vs_real}.

\begin{figure}[ht]
    \centering
    \includegraphics[width=\columnwidth]{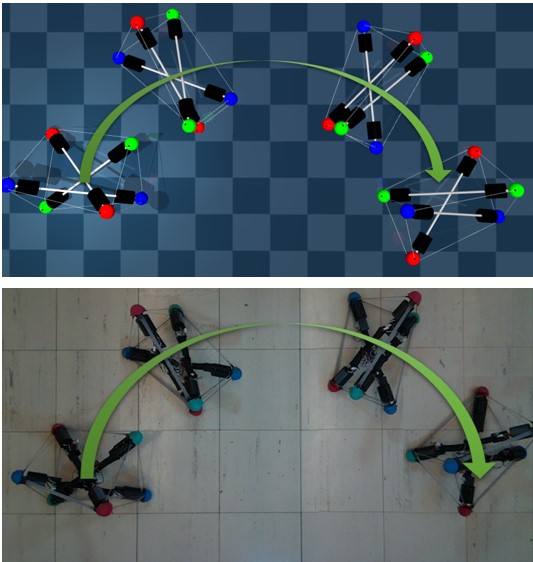}
    \vspace{-.2in}
    \caption{The frame by frame comparison of the clockwise arching trajectory execution in simulation and on the real robot.}
    \label{fig:arching_sim_vs_real}       
    \vspace{-.15in}
\end{figure}

\begin{figure}[ht]
    \centering
    \includegraphics[width=\columnwidth]{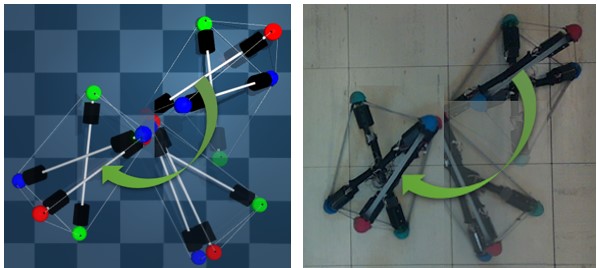}
    \vspace{-.2in}
    \caption{The frame by frame comparison of the in-place clockwise turning trajectory execution in simulation and on the real robot.}
    \label{fig:cw_sim_vs_real}       
    \vspace{-.2in}
\end{figure}

After the second iteration of system identification, two new clockwise turning polices are generated, and these policies are mapped to two long, open-loop gaits (Figure~\ref{fig:new_cw_policy}).
The deviations are larger for the arching gait (Figure~\ref{fig:new_cw_policy} Left) because of the nonuniform friction between the end caps and the ground.  This gait covers a larger area, and the coefficient of friction is not uniform everywhere on the floor.  We observe smaller deviations for the in-place turning gait (Figure~\ref{fig:new_cw_policy} Right) where the robot is confined to a much smaller region. Qualitative visualization is also avaliable in Fig~\ref{fig:arching_sim_vs_real} and Fig~\ref{fig:cw_sim_vs_real}.

\vspace{-1mm}
\section{Conclusion}


This paper demonstrates an {\tt R2S2R} pipeline that mitigates the sim2real gap for tensegrity robots and develops locomotion policies that can be seamlessly transferred to real robots.  The sim2real gap is reduced by using a differentiable physics engine that learns system parameters from real robot data via gradient descent.  The simulation can then generate new locomotion policies and extend them for long trajectories via symmetry reduction.  After transferring these policies to the real robot, the simulation can be continuously improved by re-identifying the system parameters from the recorded trajectories.  To enable efficient system identification, we introduce and experimentally validate the detachment approach for computing contact gradients and the Key Frame Loss with a trajectory segmentation strategy. In future work, the accuracy of the physics engine could be improved by allowing the system to identify the physical parameters of each component (e.g., the stiffness of each sensor tendon) and adapt online as these parameters change over time.

 


\bibliographystyle{IEEEtran}
\bibliography{references}

\begin{thebibliography}{10}
\providecommand{\url}[1]{#1}
\csname url@rmstyle\endcsname
\providecommand{\newblock}{\relax}
\providecommand{\bibinfo}[2]{#2}
\providecommand\BIBentrySTDinterwordspacing{\spaceskip=0pt\relax}
\providecommand\BIBentryALTinterwordstretchfactor{4}
\providecommand\BIBentryALTinterwordspacing{\spaceskip=\fontdimen2\font plus
\BIBentryALTinterwordstretchfactor\fontdimen3\font minus
  \fontdimen4\font\relax}
\providecommand\BIBforeignlanguage[2]{{%
\expandafter\ifx\csname l@#1\endcsname\relax
\typeout{** WARNING: IEEEtran.bst: No hyphenation pattern has been}%
\typeout{** loaded for the language `#1'. Using the pattern for}%
\typeout{** the default language instead.}%
\else
\language=\csname l@#1\endcsname
\fi
#2}}

\bibitem{lessard2016bio}
S.~Lessard, D.~Castro, W.~Asper, S.~D. Chopra, L.~B. Baltaxe-Admony,
  M.~Teodorescu, V.~SunSpiral, and A.~Agogino, ``A bio-inspired tensegrity
  manipulator with multi-dof, structurally compliant joints,'' in
  \emph{IROS}.\hskip 1em plus 0.5em minus 0.4em\relax IEEE, 2016, pp.
  5515--5520.

\bibitem{sabelhaus2018design}
A.~P. Sabelhaus, L.~J. van Vuuren, A.~Joshi, E.~Zhu, H.~J. Garnier, K.~A.
  Sover, J.~Navarro, A.~K. Agogino, and A.~M. Agogino, ``Design, simulation,
  and testing of a flexible actuated spine for quadruped robots,'' \emph{arXiv
  preprint arXiv:1804.06527}, 2018.

\bibitem{chen2020design}
M.~Chen, J.~Liu, and R.~E. Skelton, ``Design and control of tensegrity morphing
  airfoils,'' \emph{Mechanics Research Communications}, vol. 103, p. 103480,
  2020.

\bibitem{bruce2014superball}
J.~Bruce, A.~P. Sabelhaus, Y.~Chen, D.~Lu, K.~Morse, S.~Milam, K.~Caluwaerts,
  A.~M. Agogino, and V.~SunSpiral, ``Superball: Exploring tensegrities for
  planetary probes,'' \emph{12th International Symposium on Artificial
  Intelligence, Robotics, and Automation in Space (i-SAIRAS)}, 2014.

\bibitem{shah2022tensegrity}
D.~S. Shah, J.~W. Booth, R.~L. Baines, K.~Wang, M.~Vespignani, K.~Bekris, and
  R.~Kramer-Bottiglio, ``Tensegrity robotics,'' \emph{Soft Robotics}, vol.~9,
  no.~4, pp. 639--656, 2022.

\bibitem{Surovik2021AdaptiveTL}
D.~A. Surovik, K.~Wang, M.~Vespignani, J.~Bruce, and K.~E. Bekris, ``Adaptive
  tensegrity locomotion: Controlling a compliant icosahedron with
  symmetry-reduced reinforcement learning,'' \emph{The International Journal of
  Robotics Research}, vol.~40, pp. 375 -- 396, 2021.

\bibitem{wang2020end}
K.~Wang, M.~Aanjaneya, and K.~Bekris, ``Sim2sim evaluation of a novel
  data-efficient differentiable physics engine for tensegrity robots,'' in
  \emph{IEEE/RSJ International Conference on Intelligent Robots and Systems
  (IROS)}, 2021, pp. 1--8.

\bibitem{pmlr-v120-wang20b}
------, ``A first principles approach for data-efficient system identification
  of spring-rod systems via differentiable physics engines,'' in \emph{L4DC},
  vol. 120.\hskip 1em plus 0.5em minus 0.4em\relax PMLR, 10--11 Jun 2020, pp.
  651--665.

\bibitem{wang2022recurrent}
------, ``A recurrent differentiable engine for modeling tensegrity robots
  trainable with low-frequency data,'' in \emph{ICRA}, 2022.

\bibitem{bousmalis2017closing}
K.~Bousmalis and S.~Levine, ``Closing the simulation-to-reality gap for deep
  robotic learning (2019),'' \emph{Google AI Blog http://ai. googleblog.
  com/2017/10/closing-simulation-to-reality-gap-for. html}, 2019.

\bibitem{martinez2020unrealrox}
P.~Martinez-Gonzalez, S.~Oprea, A.~Garcia-Garcia, A.~Jover-Alvarez,
  S.~Orts-Escolano, and J.~Garcia-Rodriguez, ``Unrealrox: an extremely
  photorealistic virtual reality environment for robotics simulations and
  synthetic data generation,'' \emph{Virtual Reality}, vol.~24, no.~2, pp.
  271--288, 2020.

\bibitem{murthy2020gradsim}
J.~K. Murthy, M.~Macklin, F.~Golemo, V.~Voleti, L.~Petrini, M.~Weiss,
  B.~Considine, J.~Parent-L{\'e}vesque, K.~Xie, K.~Erleben, \emph{et~al.},
  ``"gradsim: Differentiable simulation for system identification and
  visuomotor control",'' in \emph{ICLR}, 2020.

\bibitem{collins2019quantifying}
J.~Collins, D.~Howard, and J.~Leitner, ``Quantifying the reality gap in robotic
  manipulation tasks,'' in \emph{2019 International Conference on Robotics and
  Automation (ICRA)}.\hskip 1em plus 0.5em minus 0.4em\relax IEEE, 2019, pp.
  6706--6712.

\bibitem{ramos2019rss}
\BIBentryALTinterwordspacing
F.~Ramos, R.~Possas, and D.~Fox, ``Bayessim: adaptive domain randomization via
  probabilistic inference for robotics simulators,'' in \emph{Robotics: Science
  and Systems (RSS)}, 2019. [Online]. Available:
  \url{https://arxiv.org/abs/1906.01728}
\BIBentrySTDinterwordspacing

\bibitem{muratore2022neural}
F.~Muratore, T.~Gruner, F.~Wiese, B.~Belousov, M.~Gienger, and J.~Peters,
  ``Neural posterior domain randomization,'' in \emph{Conference on Robot
  Learning}.\hskip 1em plus 0.5em minus 0.4em\relax PMLR, 2022, pp. 1532--1542.

\bibitem{zeng2020tossingbot}
A.~Zeng, S.~Song, J.~Lee, A.~Rodriguez, and T.~Funkhouser, ``Tossingbot:
  Learning to throw arbitrary objects with residual physics,'' \emph{IEEE
  Transactions on Robotics}, vol.~36, no.~4, pp. 1307--1319, 2020.

\bibitem{lim2021planar}
V.~Lim, H.~Huang, L.~Y. Chen, J.~Wang, J.~Ichnowski, D.~Seita, M.~Laskey, and
  K.~Goldberg, ``"real2sim2real: Self-supervised learning of physical
  single-step dynamic actions for planar robot casting",'' in
  \emph{ICRA}.\hskip 1em plus 0.5em minus 0.4em\relax IEEE Press, 2022, p.
  8282–8289.

\bibitem{sethuramantowards}
A.~Sethuraman and K.~A. Skinner, ``Towards sim2real for shipwreck detection in
  side scan sonar imagery,'' \emph{3rd Workshop on Closing the Reality Gap in
  Sim2Real Transfer for Robotics}, 2022.

\bibitem{navardi2022toward}
M.~Navardi, P.~Dixit, T.~Manjunath, N.~R. Waytowich, T.~Mohsenin, and T.~Oates,
  ``Toward real-world implementation of deep reinforcement learning for
  vision-based autonomous drone navigation with mission,'' \emph{UMBC Student
  Collection}, 2022.

\bibitem{loquercio2019deep}
A.~Loquercio, E.~Kaufmann, R.~Ranftl, A.~Dosovitskiy, V.~Koltun, and
  D.~Scaramuzza, ``Deep drone racing: From simulation to reality with domain
  randomization,'' \emph{IEEE TRO}, vol.~36, no.~1, pp. 1--14, 2019.

\bibitem{senagap}
A.~Sena, H.~Kavianirad, S.~Endo, E.~Burdet, and S.~Hirche, ``The gap in
  functional electrical stimulation simulation,'' \emph{3rd Workshop on Closing
  the Reality Gap in Sim2Real Transfer for Robotics}, 2022.

\bibitem{jabbourclosing}
J.~Jabbour, ``Closing the sim-to-real gap for ultra-low-cost,
  resource-constrained, quadruped robot platforms,'' \emph{3rd Workshop on
  Closing the Reality Gap in Sim2Real Transfer for Robotics}, 2022.

\bibitem{huang2020dynamic}
W.~Huang, X.~Huang, C.~Majidi, and M.~K. Jawed, ``Dynamic simulation of
  articulated soft robots,'' \emph{{Nature Communications}}, vol.~11, no.~1,
  pp. 1--9, 2020.

\bibitem{huang2021numerical}
X.~Huang, W.~Huang, Z.~Patterson, Z.~Ren, M.~K. Jawed, and C.~Majidi,
  ``Numerical simulation of an untethered omni-directional star-shaped swimming
  robot,'' in \emph{ICRA}.\hskip 1em plus 0.5em minus 0.4em\relax IEEE, 2021,
  pp. 11\,884--11\,890.

\bibitem{goldberg2019planar}
N.~N. Goldberg, X.~Huang, C.~Majidi, A.~Novelia, O.~M. O'Reilly, D.~A. Paley,
  and W.~L. Scott, ``On planar discrete elastic rod models for the locomotion
  of soft robots,'' \emph{{Soft Robotics}}, vol.~6, no.~5, pp. 595--610, 2019.

\bibitem{wen2022catgrasp}
B.~Wen, W.~Lian, K.~Bekris, and S.~Schaal, ``Catgrasp: Learning category-level
  task-relevant grasping in clutter from simulation,'' in \emph{ICRA}.\hskip
  1em plus 0.5em minus 0.4em\relax IEEE, 2022, pp. 6401--6408.

\bibitem{de2018end}
F.~de~Avila Belbute-Peres, K.~Smith, K.~Allen, J.~Tenenbaum, and J.~Z. Kolter,
  ``End-to-end differentiable physics for learning and control,'' in
  \emph{Advances in neural information processing systems}, 2018, pp.
  7178--7189.

\bibitem{degrave2019differentiable}
J.~Degrave, M.~Hermans, J.~Dambre, \emph{et~al.}, ``A differentiable physics
  engine for deep learning in robotics,'' \emph{Frontiers in neurorobotics},
  vol.~13, p.~6, 2019.

\bibitem{zhang2017deep}
M.~Zhang, X.~Geng, J.~Bruce, K.~Caluwaerts, M.~Vespignani, V.~SunSpiral,
  P.~Abbeel, and S.~Levine, ``Deep reinforcement learning for tensegrity robot
  locomotion,'' in \emph{ICRA}.\hskip 1em plus 0.5em minus 0.4em\relax IEEE,
  2017, pp. 634--641.

\bibitem{luo2018tensegrity}
J.~Luo, R.~Edmunds, F.~Rice, and A.~M. Agogino, ``Tensegrity robot locomotion
  under limited sensory inputs via deep reinforcement learning,'' in
  \emph{ICRA}.\hskip 1em plus 0.5em minus 0.4em\relax IEEE, 2018, pp.
  6260--6267.

\bibitem{surovik2019adaptive}
D.~Surovik, K.~Wang, M.~Vespignani, J.~Bruce, and K.~E. Bekris, ``{Adaptive
  Tensegrity Locomotion: Controlling a Compliant Icosahedron with
  Symmetry-Reduced Reinforcement Learning},'' \emph{IJRR}, 2019.

\bibitem{NTRTSim}
{NASA}, ``{NASA Tensegrity Robotics Toolkit},'' Accessed 2020,
  \url{https://github.com/NASA-Tensegrity-Robotics-Toolkit/NTRTsim}.

\bibitem{mirletz2015towards}
B.~T. Mirletz, I.-W. Park, R.~D. Quinn, and V.~SunSpiral, ``Towards bridging
  the reality gap between tensegrity simulation and robotic hardware,'' in
  \emph{IROS}.\hskip 1em plus 0.5em minus 0.4em\relax IEEE, 2015, pp.
  5357--5363.

\bibitem{caluwaerts2014design}
K.~Caluwaerts, J.~Despraz, A.~I{\c{s}}{\c{c}}en, A.~P. Sabelhaus, J.~Bruce,
  B.~Schrauwen, and V.~SunSpiral, ``Design and control of compliant tensegrity
  robots through simulation and hardware validation,'' \emph{Journal of the
  royal society interface}, vol.~11, no.~98, p. 20140520, 2014.

\bibitem{johnson2022sensor}
W.~R. Johnson, A.~Agrawala, X.~Huang, J.~Booth, and R.~Kramer-Bottiglio,
  ``Sensor tendons for soft robot shape estimation,'' in \emph{{IEEE
  Sensors}}.\hskip 1em plus 0.5em minus 0.4em\relax IEEE, 2022, pp. 1--4.

\bibitem{johnson2021integrated}
W.~R. Johnson, J.~Booth, and R.~Kramer-Bottiglio, ``Integrated sensing in
  robotic skin modules,'' in \emph{2021 IEEE Sensors}.\hskip 1em plus 0.5em
  minus 0.4em\relax IEEE, 2021, pp. 1--4.

\bibitem{lu20226ndof}
S.~Lu, W.~R. Johnson~III, K.~Wang, X.~Huang, J.~Booth, R.~Kramer-Bottiglio, and
  K.~Bekris, ``6n-dof pose tracking for tensegrity robots,'' \emph{arXiv
  preprint arXiv:2205.14764}, 2022.

\bibitem{fischler1981random}
M.~A. Fischler and R.~C. Bolles, ``Random sample consensus: a paradigm for
  model fitting with applications to image analysis and automated
  cartography,'' \emph{Communications of the ACM}, vol.~24, no.~6, pp.
  381--395, 1981.

\bibitem{surovid2018any}
D.~Surovik, J.~Bruce, K.~Wang, M.~Vespignani, and K.~E. Bekris, ``Any-axis
  tensegrity rolling via bootstrapped learning and symmetry reduction,'' in
  \emph{ISER}, Buenos Aires, Argentina, 11/2018 2018.

\bibitem{werling2021fast}
K.~Werling, D.~Omens, J.~Lee, I.~Exarchos, and C.~K. Liu, ``Fast and
  feature-complete differentiable physics engine for articulated rigid bodies
  with contact constraints,'' in \emph{RSS}, 2021.

\bibitem{detach}
{Pytorch}, ``{Pytorch Detach Method},'' Accessed 2022,
  \url{https://pytorch.org/docs/stable/generated/torch.Tensor.detach.html}.

\bibitem{heiden2022pds}
E.~Heiden, C.~E. Denniston, D.~Millard, F.~Ramos, and G.~S. Sukhatme,
  ``Probabilistic inference of simulation parameters via parallel
  differentiable simulation,'' \emph{ICRA}, 2022.

\bibitem{hu2019difftaichi}
Y.~Hu, L.~Anderson, T.-M. Li, Q.~Sun, N.~Carr, J.~Ragan-Kelley, and F.~Durand,
  ``Difftaichi: Differentiable programming for physical simulation,''
  \emph{ICLR}, 2020.

\end{thebibliography}

\clearpage

\section*{APPENDIX}

\subsection{Contact Gradients with/without "Detach"} 
To highlight our "Detach" method in the contact model, we design three toy problems to show how our method could generate correct gradient for parameter optimization.

\begin{wrapfigure}{l}{0.07\textwidth}
    \vspace{-3mm}
    \includegraphics[width=0.07\textwidth]{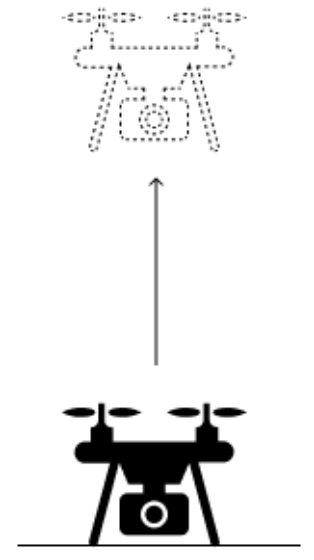}
    \label{fig:drone}
    \vspace{-10mm}
\end{wrapfigure}
The first test evaluates  Algorithm~\ref{alg:detach_restitution} and corresponds to a trajectory optimization challenge from the literature~\cite{werling2021fast}, where a drone is taking off from the ground and reaching a fixed height at $t=500$. The loss is the Mean Square Error (MSE) of the estimated drone's height $\hat{x}$ against the ground truth height that is $x=10m$.

\begin{figure}[!htpb]
    \centering
    \includegraphics[width=0.31\columnwidth]{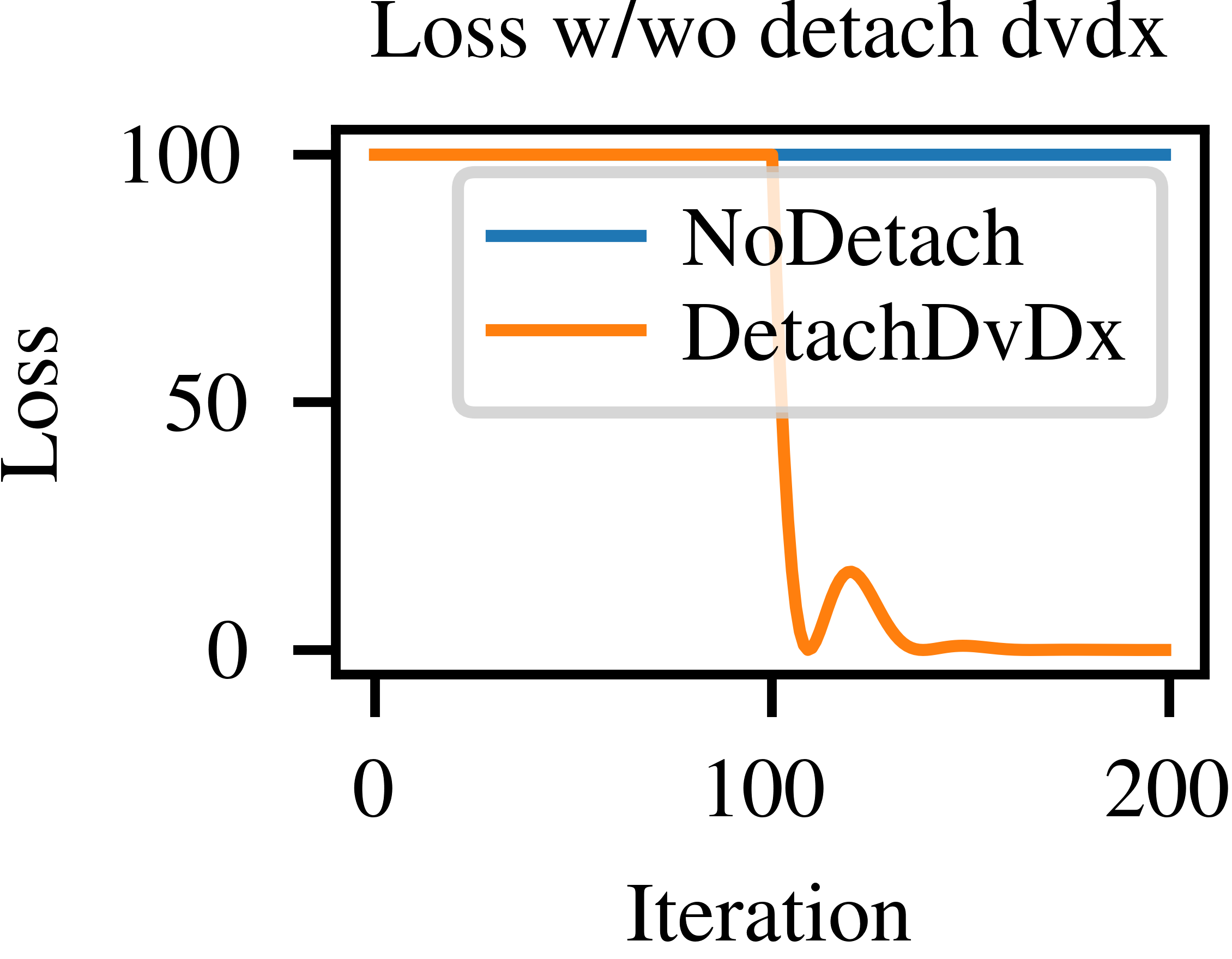}
    \includegraphics[width=0.31\columnwidth]{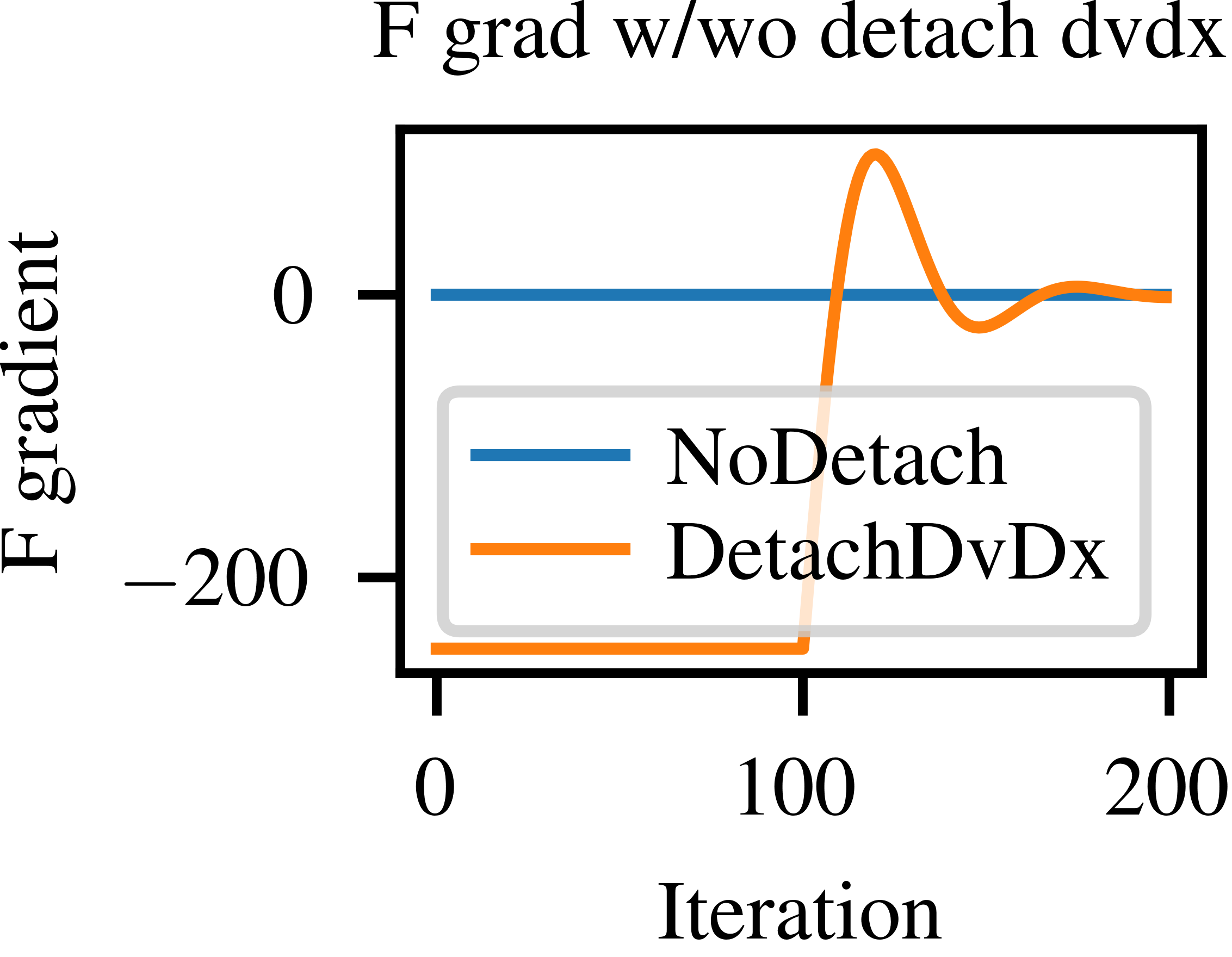}
    \includegraphics[width=0.31\columnwidth]{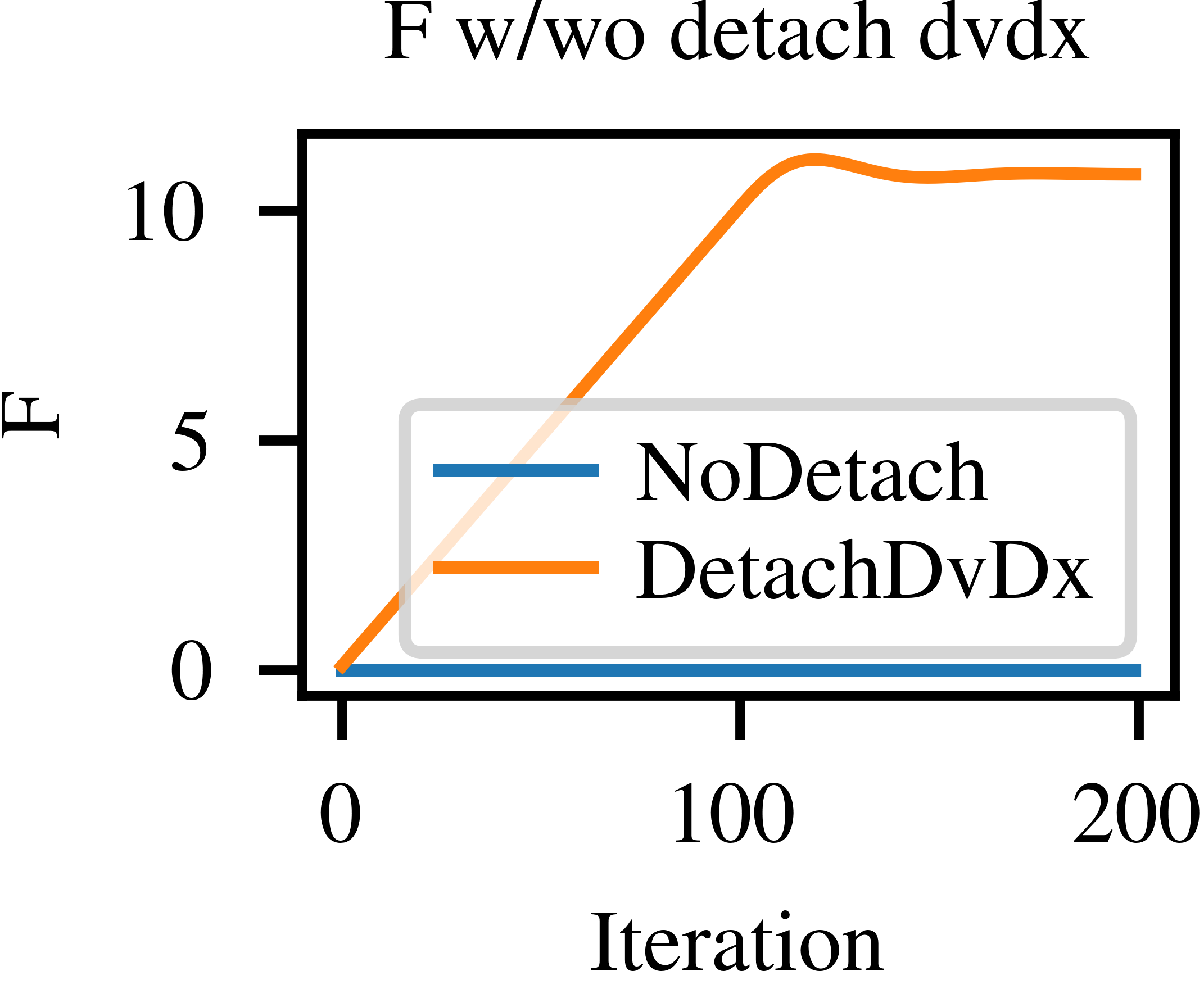}
    \caption{We train a drone to lift off the ground to evaluate our "Detaching" method. Loss is the MSE of the distance from drone $\hat{x}$ to target $x$ at $t=500$. The drone is initialized resting on the ground. The gradient $\dfrac{\partial \hat{x}}{\partial F}$ is zero with clamping contact. However, after detaching the  contact response $\Delta v$(dv) from the computation graph, the correct gradient on $F$ can guide the optimizer to increase $F$ and reach the target.}
    \label{fig:drone_lift}
\end{figure}

The clamping contact between the drone and ground causes zero gradients and halts progress in the optimization as the passive contact generates an opposite equal gradient to F. After detaching the contact velocity impulse $\Delta v$ and position impulse $\Delta x$ from the computation graph of Algorithm~\ref{alg:detach_restitution}, the correct gradient keeps increasing $F$ even if there is no change in the loss for the first 100 iterations (Figure~\ref{fig:drone_lift}).

\begin{wrapfigure}{l}{0.2\textwidth}
    \vspace{-3mm}
    \includegraphics[width=0.2\textwidth]{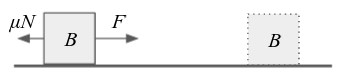}
    \label{fig:box}
    \vspace{-8mm}
\end{wrapfigure} We bring the box problem, pulling a box to the target position in 200 time steps, to evaluate Algorithm~\ref{alg:detach_mu}. The pulling force $F$ is known and the ground friction coefficient $\mu$ is to be estimated. We take the MSE of box position and target position as a loss function.

The initial $\mu$ is 1 and box can't be moved with such large friction. Our detach method can generate correct negative gradient to reduce $\mu$ and the loss. However, without our method, the gradient is always zero and the loss curve never goes down.

\begin{figure}[!htpb]
    \centering
    \includegraphics[width=0.31\columnwidth]{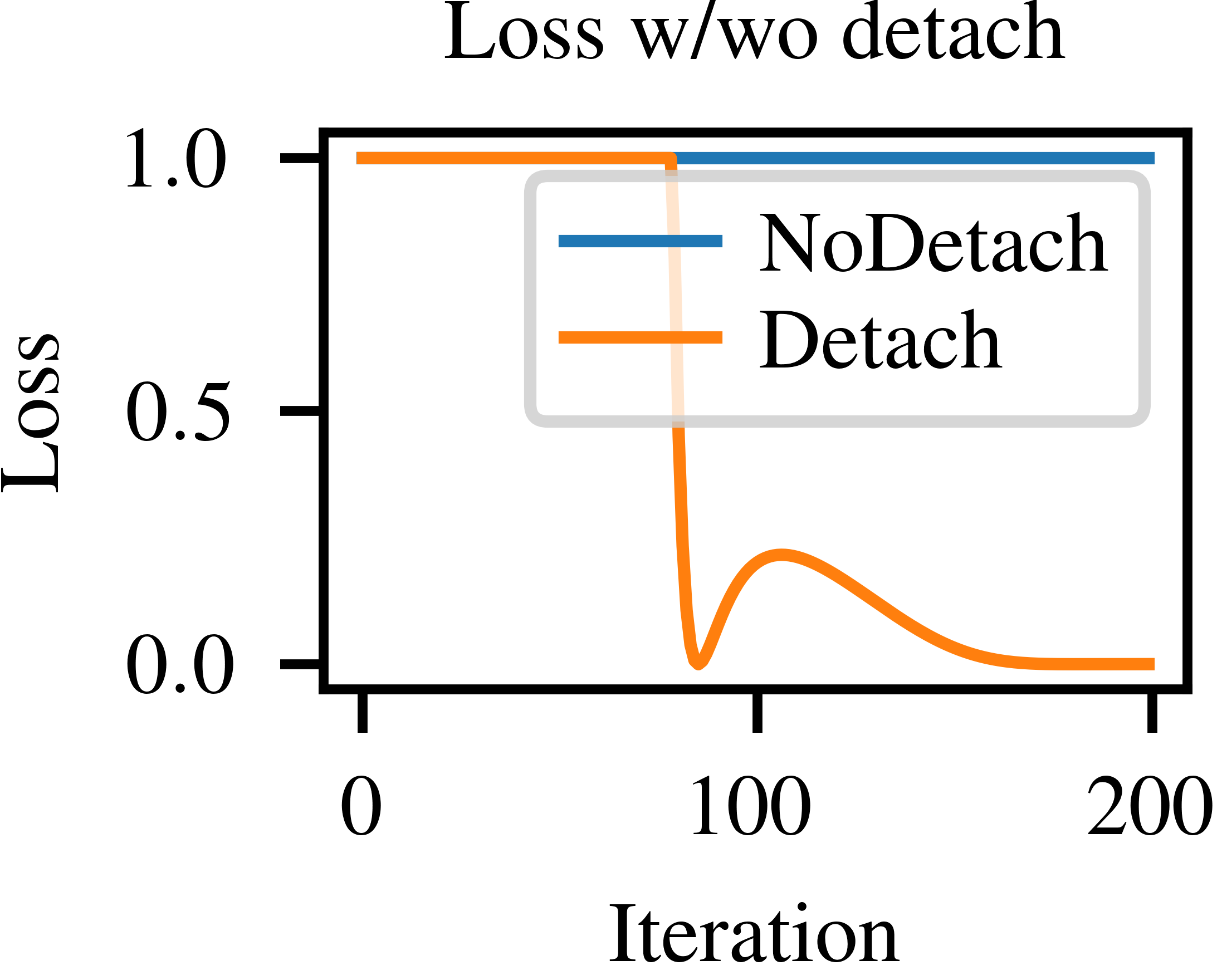}
    \includegraphics[width=0.31\columnwidth]{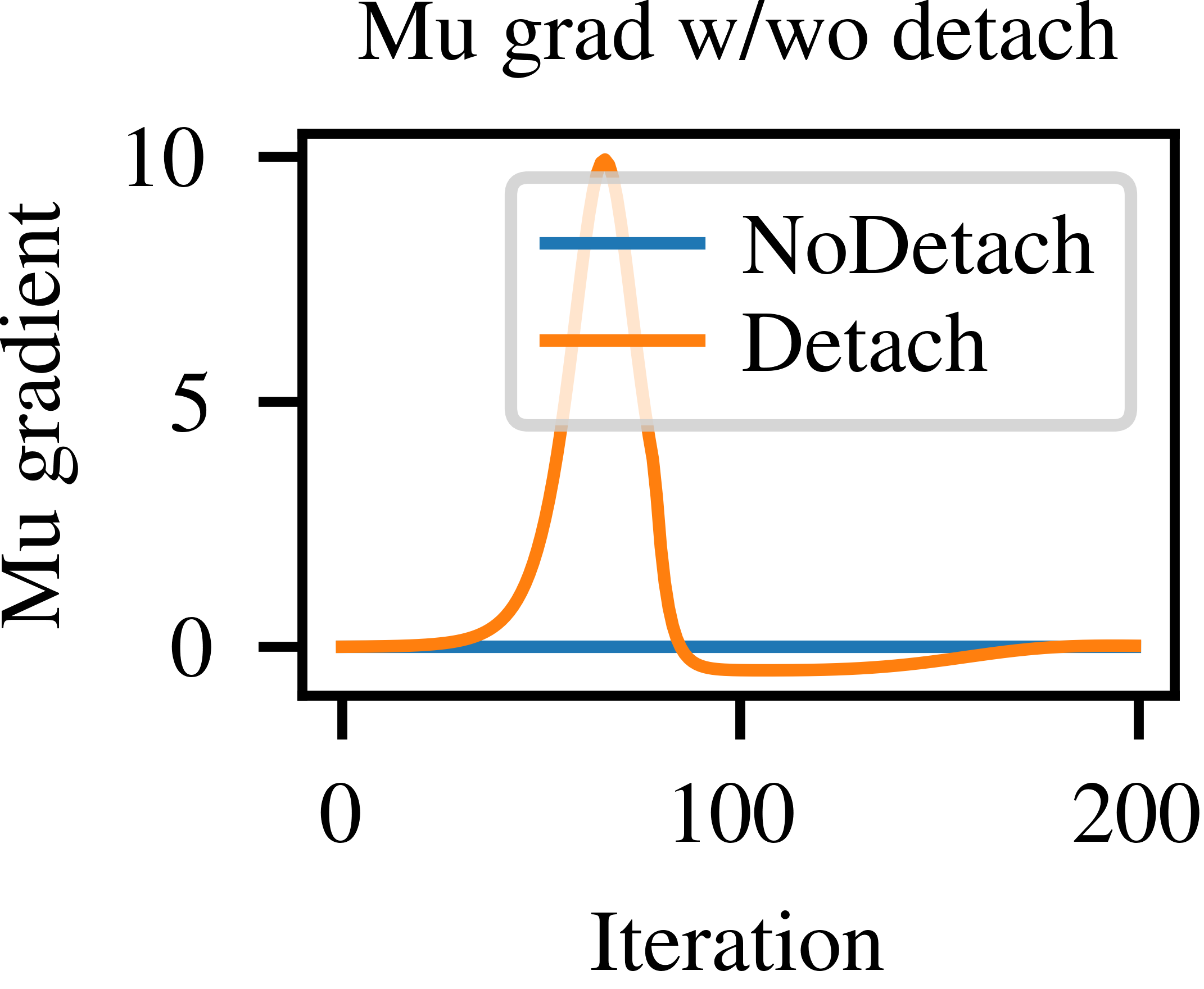}
    \includegraphics[width=0.31\columnwidth]{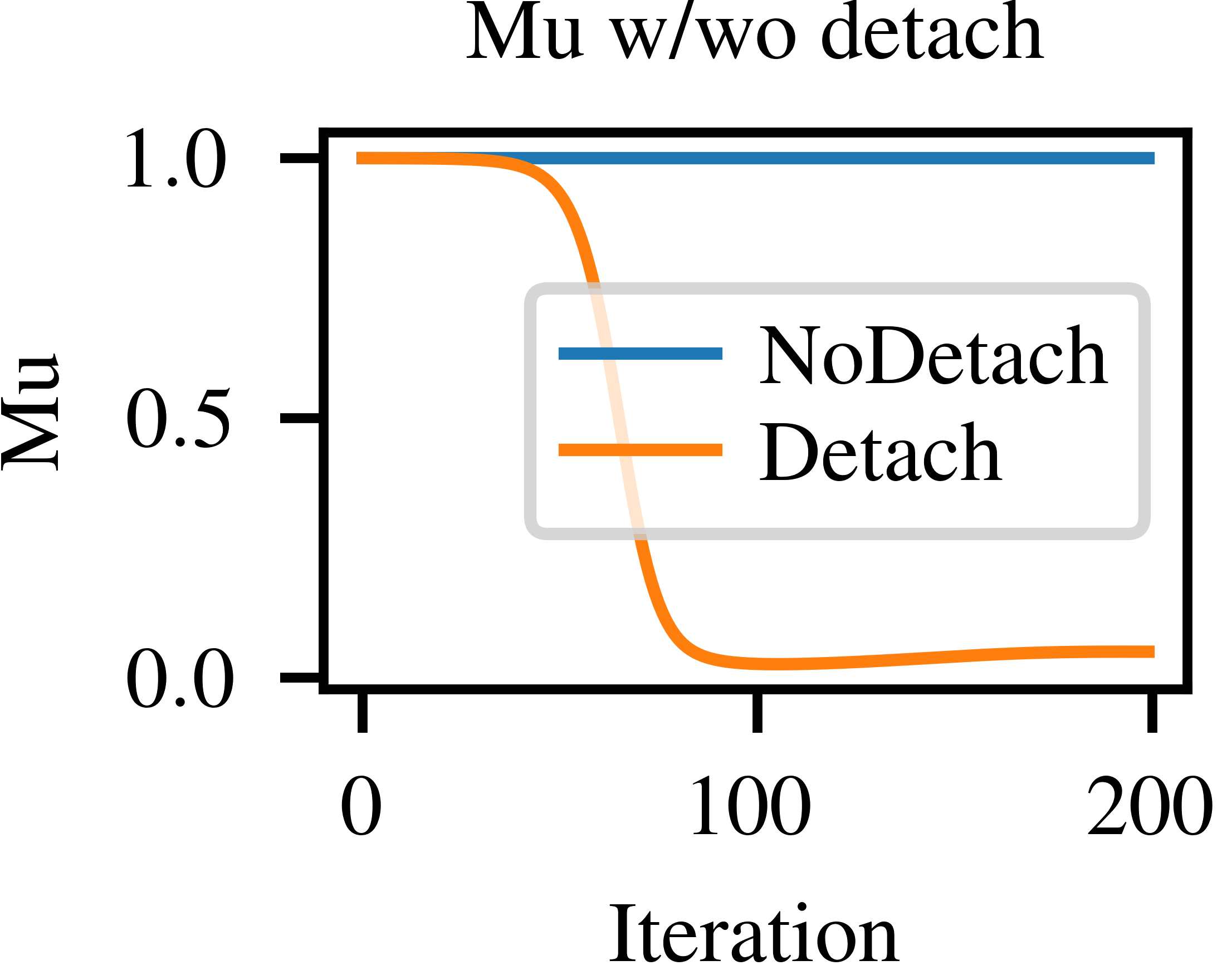}
    \caption{The box pulling problem shows the detaching method can get correct gradient to reduce the friction coefficient $\mu$. Otherwise, zero gradients would happen on optimization of $\mu$.}
    \label{fig:box_result}
\end{figure}

In general, we suggest to apply the "detach" method to clamping contacts of all actuated objects. However, this method should not be applied to objects whose movement only relies on these passive contact forces, e.g. the billiard. The philosophy is that we keep the "trending" gradient from the small actuation force even if the object can't move. This avoids the gradient discontinuities between clamping contact and separating contact.

\begin{wrapfigure}{l}{0.2\textwidth}
    \vspace{-3mm}
    \includegraphics[width=0.2\textwidth]{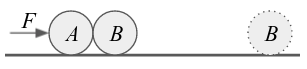}
    \label{fig:billiards}
    \vspace{-8mm}
\end{wrapfigure} Finally, we take the billiards problem, applying $F$ on ball $A$ and aiming the ball $B$ to reach a target position in 500 timesteps, to evaluate the detaching philosophy - only detach actuated object. The loss is the mean MSE of $B$'s position and target position. 

\begin{figure}[!htpb]
    \centering
    \includegraphics[width=0.31\columnwidth]{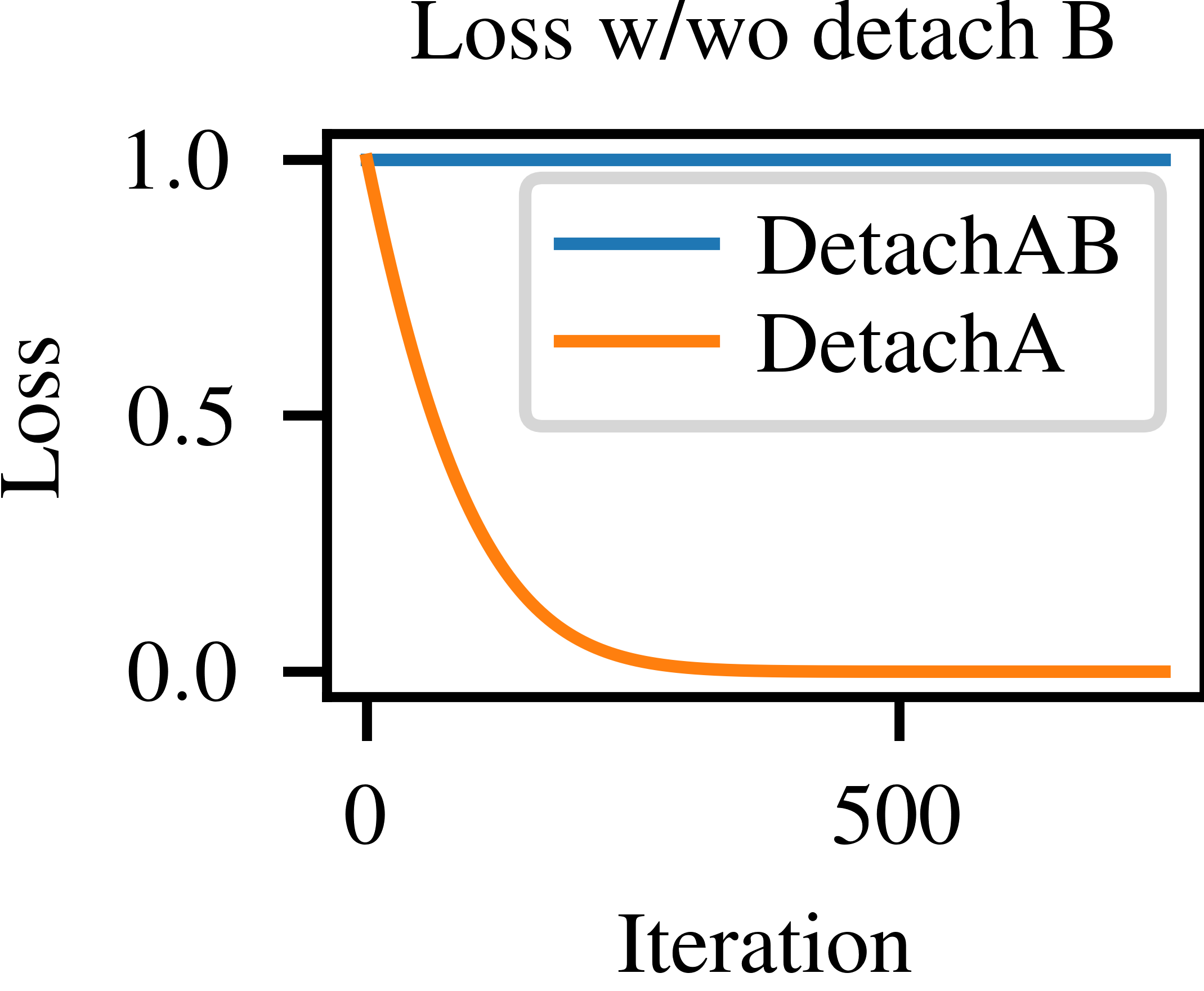}
    \includegraphics[width=0.31\columnwidth]{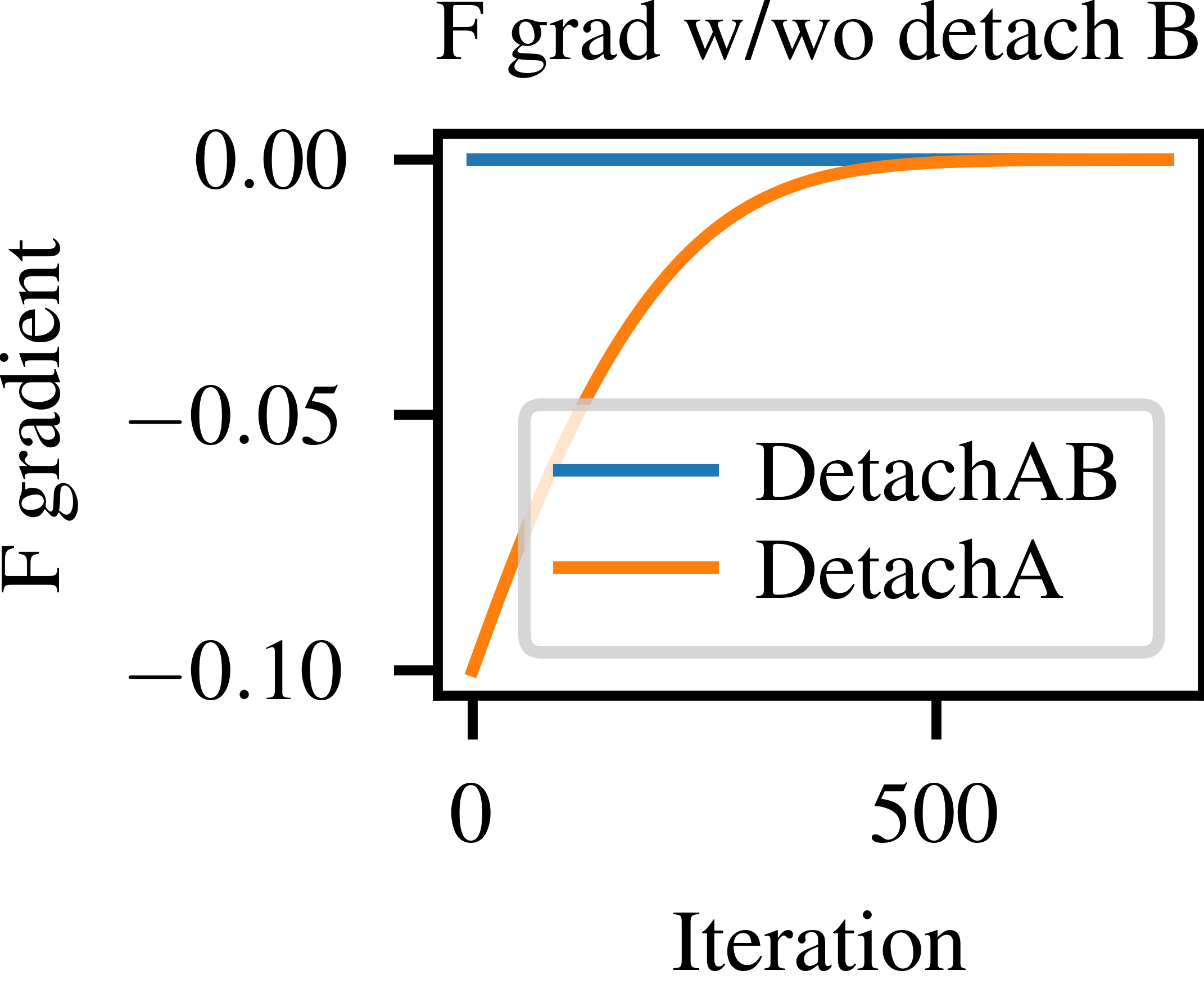}
    \includegraphics[width=0.31\columnwidth]{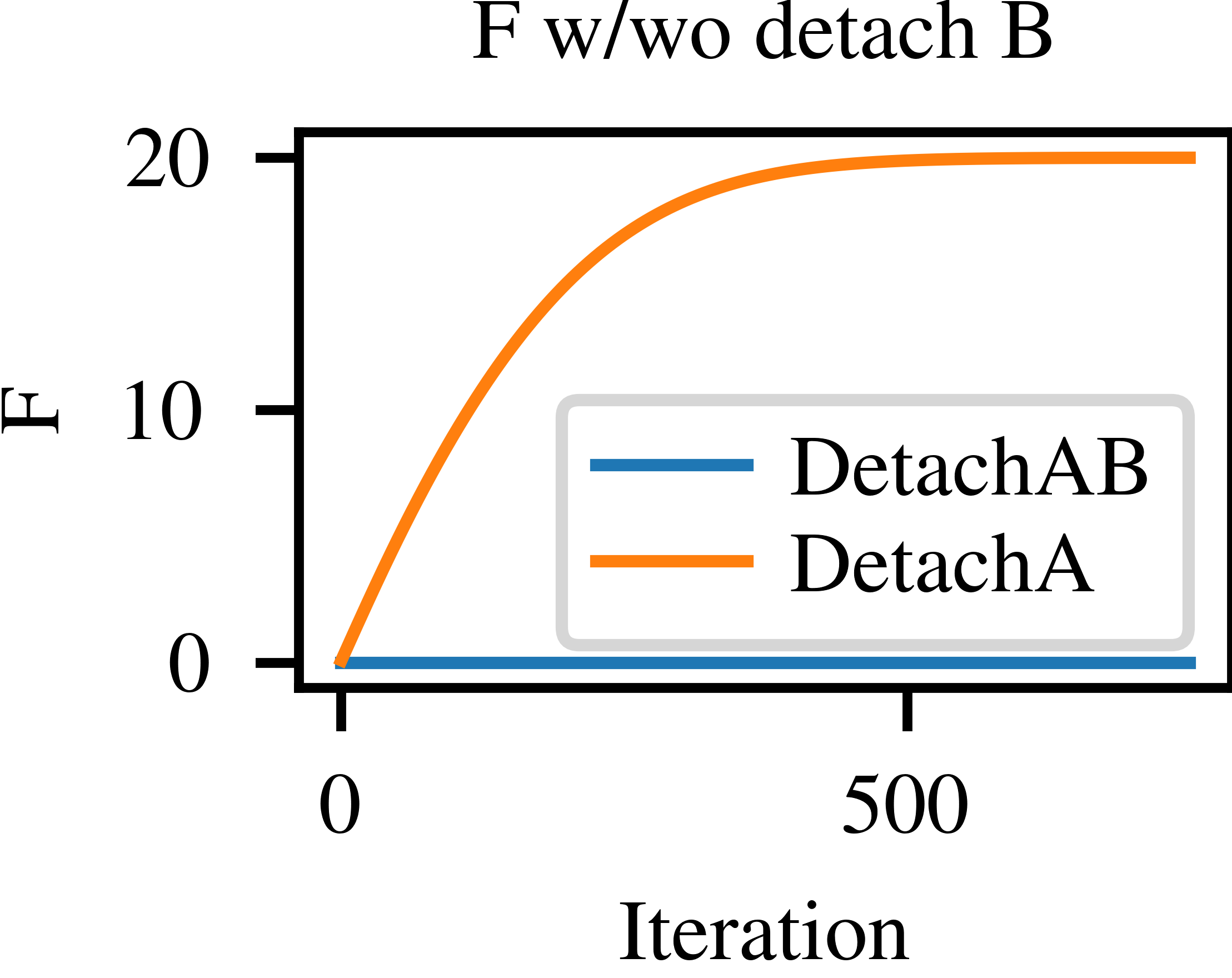}
    \caption{The billiards problem shows that only detaching the actuated object $A$ can get the correct gradient on actuation force $F$ to reduce the loss. Detaching the passive moved object $B$ will isolate $B$ from gradient passing, which result to zero gradients on $F$.}
    \label{fig:billiards_result}
\end{figure}

We compare the differences between detaching both $A, B$ and detaching $A$ only. As we suggested, we should only detach the actuated object $A$'s contact response. We shouldn't detach $B$'s contact response, Because $B$ is only affected by the passive contact forces. The result in Figure~\ref{fig:billiards_result} shows that detaching both A and B could get zero gradient on $F$ and no change in $F$ and loss. However, if we only detach the actuated object $A$, we can still get the correct gradient direction and escape the saddle point.

\subsection{Gaits Transition Smoothing}
After generating locomotion gaits for all bottom triangles, we have a controller to generate longer trajectories. However, these trajectories have redundant actuation between gait transitions. We adopt a smoothing algorithm to smooth out the transition.

\begin{figure}[!htpb]
    \centering
    \includegraphics[width=\columnwidth]{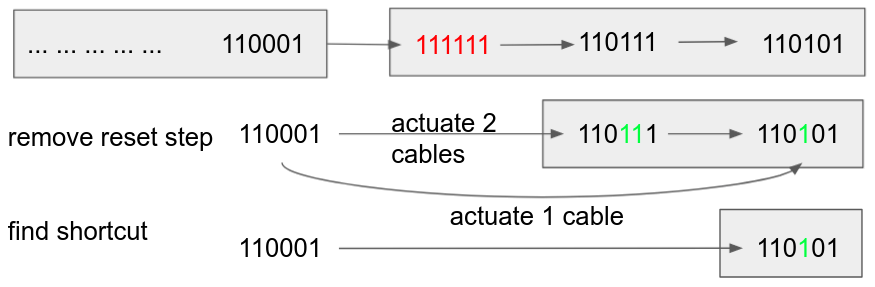}
    \caption{The current gait ends at state 110001, i.e., cables 1,2,6 are at maximum length and cables 3,4,5 are at minimum length. The next gait includes three steps. Step one, which is to reset the robot to a neural state, is removed. Then we compute the distance from 110001 to the remaining two steps. We transit to step 3 directly since fewer cables are actuated.}
    \label{fig:gait_transition_smoothing}
\end{figure}

Let's consider a simple example of gaits transition in Figure~\ref{fig:gait_transition_smoothing}. Although the following gait includes 3 steps, we only keep the third step after removing the reset step and finding the shortcut with minimum actuated cables. But these shortcuts should be reevaluated because some of them may not work due to changes in center of gravity.

\subsection{Comparison to Alternatives on a Small-scale Problem}
\label{sec:small_scale_problem}

This section evaluates Key Frame Loss (KFL) on a small-scale problem where the MS alternative can be applied. A learning rate of 0.1 and an Adam optimizer are used.

\begin{wrapfigure}{r}{0.2\textwidth}
    \vspace{-.2in}
    \includegraphics[width=0.2\textwidth]{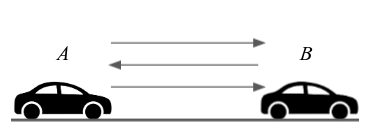}
    \label{fig:car_driving}
    \vspace{-8mm}
\end{wrapfigure}
\begin{figure}[!htpb]
    \vspace{-.15in}
    \centering
    \includegraphics[width=0.49\columnwidth]{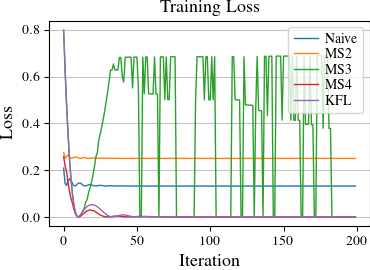}
    \includegraphics[width=0.49\columnwidth]{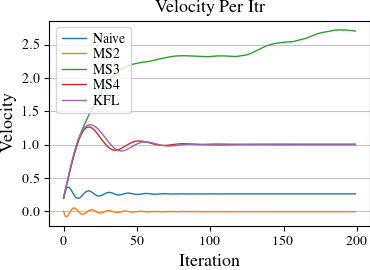}
    \vspace{-.2in}
    \caption{Loss and velocity curve comparison during the training process of Naive, multiple shooting with 2, 3, and 4 windows (MS), and the Key Frame Loss method (KFL). 
    }
    \label{fig:kfl_eval}
    \vspace{-4mm}
\end{figure}

The problem involves identifying the velocity of a vehicular model so as to match a demonstration trajectory from a ground truth vehicle that drives between points A and B. The ground truth trajectory uses a velocity $v=1$~m/s to reach point B at $t=100$, return to A at $t=200$, and reach B again at $t=300$.  We compare 3 methods: a) A naive approach that computes loss at each step as the difference between ground truth and estimated trajectory; b) A multiple shooting (MS) method~\cite{heiden2022pds} that splits the trajectory to multiple windows and computes the loss by defects at the end of each window (different numbers of windows are tested for variants MS2, MS3, MS4); and c) the proposed KFL method. Fig.~\ref{fig:kfl_eval} provides the comparison.

Only KFL and MS4 converge to the correct velocity (1~m/s). The other methods fail due to conflicting gradients. We also tested MS for a higher number of windows (5 to 9). The resulting loss curves are non-smooth, but the velocity eventually converges. The MS method samples discrete gradients on a continuous trajectory, so the sampling frequency must be high enough to compute correct gradients, which need to be discovered empirically. At the same time, the frequency should not be too high for noisy ground truth data, as the noise may dominate the optimization. The KFL approach splits the trajectory by ``gaits'' (i.e., the trajectory segment where the robot moves to a specific point), and this allows it to compute correct gradients.

\subsection{\textcolor{black}{Detachment Method for Restitution with Time of Impact (TOI)}}
For more accurate contact point computation, the Time of Impact (TOI) is introduced ~\cite{hu2019difftaichi} to compute the exact time that the contact happens. The detachment method can also integrate with TOI for restitution inference, which is shown in Algorithm~\ref{alg:detach_restitution_toi}.

\subsection{\textcolor{black}{Tensegrity Robot Locomotion and Applications}}

Tensegrity robots' compliance gives them the ability to adapt to unstructured terrain and survive harsh impacts, motivating them as future planetary rovers~\cite{caluwaerts2014design}.  Many tensegrity robots achieve locomotion by changing their tendon lengths to shift their center of mass outside of their polygon of stability and therefore roll~\cite{shah2022tensegrity}.  Electric motors that drive winches are commonly used to extend and contract tensegrity robots' cables~\cite{bruce2014superball}.  The 3-bar tensegrity robot used in this work has six such motors that drive winches to extend and contract six cables (the remaining three tendons are passive elastic elements) in order to shift the robot's center of mass and achieve locomotion.  The robot can demonstrate the different locomotion policies enumerated in this paper (forward rolling, backward rolling, counterclockwise turning, and clockwise turning) by controlling its six cable lengths to match a sequence of target shapes.  Sensor tendons~\cite{johnson2022sensor} that run parallel to each actuator sense the length of the cables and provide feedback to a PID controller so that the robot can execute the locomotion policies with high fidelity. In future work, we aim to demonstrate this lightweight, adaptable tensegrity robot navigating unstructured terrain and maintaining robust control even when subjected to harsh impacts.

\vspace{-3mm}
\begin{algorithm}
\caption{Detachment Method for Restitution with TOI}\label{alg:detach_restitution_toi}
\begin{algorithmic}
\State $v_{t+1}' = v_t - g \Delta t + F/m\Delta t$
\State {$x_{t+1}' = x_t + v_{t+1}' \Delta t$}
\State $\Delta v = -(1+e)v_{t+1}'.detach()$
\State $v_{t+1} = v_{t+1}' + \Delta v$
\If { $x_{t} > ground$}
    \State {$toi = (x_{t} - ground) / max(-v_{t}, 10^{-4})$}
    \State {$toi = toi.detach()$}
\Else
    \State {$toi = 0$}
\EndIf
\If { $x_{t+1}' < ground$}
    \State {$\Delta x =  ground-x_{t+1}'$}
    \State {$x_{t+1} = x_t + v_{t+1} \Delta t + \Delta x.detach() - toi \Delta v$}
\Else
    \State {$x_{t+1} = x_t + v_{t+1} \Delta t$}
\EndIf
\end{algorithmic}
\end{algorithm}
\vspace{-3mm}

\end{document}